\documentclass[10pt,journal,compsoc]{IEEEtran}

\ifCLASSOPTIONcompsoc
  \usepackage[nocompress]{cite}
\else
  \usepackage{cite}
\fi
\ifCLASSINFOpdf
\else
\fi

\usepackage{graphicx}
\usepackage{amsfonts,amssymb}
\usepackage{amsmath}
\usepackage{url}
\usepackage{multirow}
\usepackage{float}
\usepackage{subfigure}
\usepackage{todonotes}
\usepackage{enumitem}
\usepackage{algorithmicx}
\usepackage{algorithm}

\usepackage{algpseudocode}
\usepackage{amsmath}

\begin{document}

\newcommand{\ie}{\emph{i.e., }}
\newcommand{\eg}{\emph{e.g., }}
\newcommand{\etal}{\emph{et al.}}
\newcommand{\st}{\emph{s.t. }}
\newcommand{\etc}{\emph{etc.}}
\newcommand{\wrt}{\emph{w.r.t. }}
\newcommand{\cf}{\emph{cf. }}
\newcommand{\aka}{\emph{aka. }}

\title{RCC-Dual-GAN: An Efficient Approach for Outlier Detection with Few Identified Anomalies}

\author{Zhe~Li, Chunhua~Sun, Chunli~Liu, Xiayu~Chen, Meng~Wang and Yezheng~Liu
\thanks{Yezheng Liu is the corresponding author.}
\IEEEcompsocitemizethanks{
	\IEEEcompsocthanksitem Z. Li, C. Sun, C. Liu, X. Chen, and Y. Liu are with School of Management, Hefei University of Technology, China. E-mail: lizhe@mail.hfut.edu.cn and \{sunchunhua, liuchunli, xychen, liuyezheng\}@hfut.edu.cn.
	\IEEEcompsocthanksitem M. Wang is with School of Computer Science and Information Engineering, Hefei University of Technology, China.
E-mail: eric.mengwang@gmail.com.}
}

\markboth{}
{}

\IEEEtitleabstractindextext{
\begin{abstract}

Outlier detection is an important task in data mining and many technologies have been explored in various applications. However, due to the default assumption that outliers are non-concentrated, unsupervised outlier detection may not correctly detect group anomalies with higher density levels. As for the supervised outlier detection, although high detection rates and optimal parameters can usually be achieved, obtaining sufficient and correct labels is a time-consuming task. To address these issues, we focus on semi-supervised outlier detection with few identified anomalies, in the hope of using limited labels to achieve high detection accuracy. First, we propose a novel detection model Dual-GAN, which can directly utilize the potential information in identified anomalies to detect discrete outliers and partially identified group anomalies simultaneously. And then, considering the instances with similar output values may not all be similar in a complex data structure, we replace the two MO-GAN components in Dual-GAN with the combination of RCC and M-GAN (RCC-Dual-GAN). In addition, to deal with the evaluation of Nash equilibrium and the selection of optimal model, two evaluation indicators are created and introduced into the two models to make the detection process more intelligent. Extensive experiments on both benchmark datasets and two practical tasks demonstrate that our proposed approaches (\ie Dual-GAN and RCC-Dual-GAN) can significantly improve the accuracy of outlier detection even with only a few identified anomalies. Moreover, compared with the two MO-GAN components in Dual-GAN, the network structure combining RCC and M-GAN has greater stability in various situations. The experiment codes are available at: \url{https://github.com/leibinghe/RCC-Dual-GAN}.
\end{abstract}

\begin{IEEEkeywords}
Semi-supervised outlier detection, few identified anomalies, group anomalies,Dual-GAN, robust continuous clustering
\end{IEEEkeywords}}

\maketitle

\IEEEdisplaynontitleabstractindextext
\IEEEpeerreviewmaketitle

\section{Introduction}

\IEEEPARstart{O}{utliers} refer to observations that have significantly different characteristics from the majority of other data. These observations are so unique as to arouse suspicions that they were generated by illegal acts or undetected errors. To reveal the critical and interesting information in them, many outlier detection technologies have been studied and applied in various applications. Such as the fraud detection in credit card transaction~\cite{Fiore2017Using,Pozzolo2017Credit,Makki2019An}, fake rating and review detection in e-commerce service platform~\cite{Rahman2017Search,Liu2019A}, intrusion detection in network service request~\cite{Preeti2019A,Raman2017A}, and abnormal moving object detection in traffic monitoring~\cite{Mao2017Feature}.

In general, according to the availability of data labels, existing methods can be divided into three categories: unsupervised, supervised, and semi-supervised outlier detection. Unsupervised algorithms are among the most widely studied because they do not require additional labels or prior information. Including statistical-based~\cite{Yang2009Outlier,Bo2018Deep}, cluster-based~\cite{Manzoor2016Fast}, regression-based~\cite{Paulheim2015A}, proximity-based~\cite{Salehi2016Fast,Chehreghani2016K,Makki2019Adapted}, reconstruction-based~\cite{Zhou2017Anomaly,Schlegl2017Unsupervised,Adversarially2018Sabokrou}, and other approaches. They assume explicitly or implicitly that outliers are not as concentrated as normal data~\cite{Steinwart2005A}. Thus, discrete anomalies can be detected effectively. However, in many cases, multiple anomalies (\eg DoS attack) may be generated by the same mechanism. They become increasingly concentrated such that unsupervised outlier detection incorrectly detects these group anomalies as normal data. Moreover, the selection of models and parameters is a considerable challenge for unsupervised methods without the help of prior knowledge. As for supervised algorithms, higher detection rates and optimal parameters can usually be obtained because the labels are complete and correct during training~\cite{Zhang2018Anomaly}. However, obtaining sufficient anomalies and correct labels is a time-consuming task. In addition, detection models trained on fully labeled data have considerable uncertainty when dealing with emerging anomalies.

To address these issues, semi-supervised outlier detection with few identified anomalies and abundant unlabeled data was proposed~\cite{Aggarwal2017Outlier}. Despite insufficient capacity to label all normal examples or outliers, few abnormal behaviors that have triggered an alarm can be collected easily in many applications~\cite{Zhang2018Anomaly}. Examples include DoS attacks that have caused a system crash and insurance applications that have been proven to be fraudulent. In addition to their own labels, these identified anomalies can also provide a priori information for other samples that have the same generation mechanism. If this information is utilized fully, semi-supervised model can not only identify discrete anomalies, but also detect partially identified group anomalies. Moreover, few anomalies can also provide valuable guidance for the selection of models and parameters, which has significant advantage over unsupervised outlier detection. Thus, this paper will focus on this special anomaly detection setting, in the hope of using limited tags to achieve high detection accuracy. 

The initial model~\cite{Daneshpazhouh2014Entropy,Daneshpazhouh2014Semi}, first extracts reliable normal examples through a heuristic method, which is completely consistent with the first step of PU-learning. Then a modified outlier detection model is trained on the new tagged dataset to identify the other anomalies. But since the outliers are usually discrete or belong to different clusters, the extracted samples that are significantly different from identified anomalies are not necessarily normal. As a result, the potential information in identified anomalies is not used effectively and erroneous information may also be introduced into the new tagged dataset. Therefore, to augment the use of known information and reduce the introduction of error messages, several soft versions of the above strategy were established. For example, LBS-SVDD~\cite{Bo2014An} assigns abnormal likelihood values to each sample based on the proportion of anomalies in its neighbors, while ADOA~\cite{Zhang2018Anomaly} attaches a weight to each instance according to its own isolation and its similarity to identified anomalies. However, the calculation of neighbors and similarity usually has a high computational cost and is likely to be affected by irrelevant variables.

In this paper, we propose a one-step method for semi-supervised outlier detection with few identified anomalies, which can directly utilize the potential information in identified anomalies without calculating the abnormal degree of each instance. Specifically, the Dual Generative Adversarial Networks (Dual-GAN) contains two Multiple-Objective Generative Adversarial Networks~\cite{Liu2019Generative} (\ie UMO-GAN and AMO-GAN) and an overall discriminator. The Unlabeled MO-GAN (UMO-GAN) is used to learn the generation mechanism of unlabeled data and gradually generates informative potential outliers to provide a reasonable reference distribution for unlabeled data. By contrast, the Abnormal MO-GAN (AMO-GAN) is used to learn the deep representation of identified anomalies and generates numerous potential anomalies with the same generation mechanism as known anomalies to enhance the minority class. Thus, in order to distinguish these identified and synthesized anomalies from the unlabeled data, the overall discriminator will not only describe a division boundary that encloses the concentrated data, but will also separate partially identified group anomalies from the concentrated data. In addition, considering that instances with similar output values are not necessarily close to one another in the sample space, we replace the MO-GAN with Multiple Generative Adversarial Networks (M-GAN). More specifically, the modified model RCC-Dual-GAN first divides the identified anomalies and unlabeled data into different subsets through a Robust Continuous Clustering (RCC)~\cite{Shah2017Robust}. Then, multiple GANs are utilized to learn their generation mechanisms directly. Compared with the original model Dual-GAN, RCC-Dual-GAN can create the reference distribution and augment the minority class more robustly in various situations. The main contributions of this work are summarized as follows: 

\begin{itemize}
\item We propose a semi-supervised outlier detection method Dual-GAN, which consists of two MO-GAN and an overall discriminator. The method utilizes the potential information in identified anomalies directly to detect discrete anomalies and partially identified group anomalies simultaneously. 

\item Considering that instances with similar output values may not all be similar in a complex data structure, we change the original model Dual-GAN to RCC-Dual-GAN by replacing MO-GAN with the combination of RCC and M-GAN. Compared with Dual-GAN, the modified model can create the reference distribution and augment the minority class more robustly.

\item Considering the difficulty in finding the Nash equilibrium and optimal model during iteration, two evaluation indicators are created and introduced into the two models to make the detection process more intelligent.

\item We conduct extensive experiments on both benchmark datasets and two practical tasks to investigate the performance of our proposed approaches. The results show that even with only a few identified anomalies, our proposed approaches can significantly improve the accuracy of outlier detection. 
\end{itemize}

The rest of this paper is organized as follows. Section \ref{sec:related} provides a brief review of related works. Section \ref{sec:Background} introduces the detection principle and model details of MO-GAN, and the proposed models are described in Section \ref{sec:Semi}. We report extensive experiment results in Section \ref{sec:4} and the whole paper is concluded in Section \ref{sec:conclusions}.


\section{Related Work}
\label{sec:related}
Numerous overviews on outlier detection algorithms for different kinds of data and applications are available in the literature~\cite{Aggarwal2017Outlier,Wang2019Progress}. Here, we briefly discuss the common outlier detection methods (\ie unsupervised and supervised approaches) and then focus on the semi-supervised outlier detection with limited labels, which is most relevant to our research. Finally, the GAN-based outlier detection algorithms are reviewed in Section \ref{sec:GAN}.

\subsection{Common Outlier Detection Methods}
\label{sec:Common}

Unsupervised outlier detection methods have been studied widely because they require no additional label. Specific algorithms include proximity-~\cite{Salehi2016Fast,Chehreghani2016K,Makki2019Adapted}, statistical-~\cite{Yang2009Outlier,Bo2018Deep}, cluster-~\cite{Manzoor2016Fast}, regression-~\cite{Paulheim2015A}, and reconstruction-based models~\cite{Zhou2017Anomaly,Schlegl2017Unsupervised,Adversarially2018Sabokrou}. Proximity-based models assume the outliers are points far away from other data and can be performed by measuring the distance or density of the point. By contrast, the remaining models assume that outliers are observations that have large deviations from the normal profiles and can be performed by creating a model for the majority of samples. However, all these algorithms based on the assumption that outliers are not as concentrated as the normal data, such that the group anomalies with higher density levels cannot be detected correctly. And most of them must be provided with model assumptions or parameters in advance, which is a huge challenge for unsupervised methods without the help of prior knowledge.

Supervised outlier detection can be considered as a special classification problem and many classification algorithms have been applied. However, in most practical applications, outliers are far less common than normal data, so that the direct use of off-the-shelf classifiers may produce biased results. Hence, cost-sensitive learning~\cite{Wang2010Boosting} and adaptive re-sampling~\cite{Fiore2017Using,Lima2017Feature,Lima2019Heartbeat}, are later incorporated into the classification process. The cost-sensitive learning increases the misclassification costs of outliers by weighting the classification errors, whereas the adaptive re-sampling increases the relative proportion of the minority class by under- or over-sampling. Supervised algorithms usually achieve good parameters and high detection rates because the labels are complete during training. But the question is how to obtain sufficient anomalies and correct labels, which is a time-consuming and expensive task. Moreover, the detection model trained on the fully labeled data has significant uncertainty in dealing with emerging anomalies.

\subsection{Semi-Supervised Outlier Detection Methods}
\label{sec:semi}

According to the available labels, semi-supervised outlier detection can be divided into three categories: one-class learning with only normal examples, semi-supervised outlier detection with small amount of labeled data, and semi-supervised outlier detection with few identified anomalies. The one-class learning is only slightly different from unsupervised outlier detection, and most of the unsupervised approaches (\eg OC-SVM~\cite{Erfani2016High} and SVDD~\cite{Liu2013SVDD}) can be used in this case~\cite{Aggarwal2017Outlier}. The outlier detection model established on the one-class dataset tends to be more robust because of the absence of additional interference from anomalies. However, considerable time must be spent in verifying the collected samples to ensure that the training data contain only normal data. The semi-supervised outlier detection with small amount of labeled data usually optimizes an outlier detection model (\eg $k$-means ~\cite{Gao2006Semi} and fuzzy rough $k$-means~\cite{Xue2010Semi}) with the assurance that the labels of the labeled data are almost unchanged. Compared with unsupervised models, their performance is improved through a small amount of labeled data. However, the potential information in the labeled examples is not used effectively. And the normal examples in the labeled data may still require additional confirmation because of undetected anomalies.

Compared with them, the case of semi-supervised outlier detection with few identified anomalies is much simpler because few abnormal behaviors can be easily collected in many applications. The initial model~\cite{Daneshpazhouh2014Entropy,Daneshpazhouh2014Semi} first extracts reliable normal examples through a heuristic method, and then trains a semi-supervised outlier detection model described above on the new labeled dataset. But since the outliers are usually discrete or belong to different clusters, the extracted samples that are far from identified anomalies are not necessarily normal. As a result, the potential information in identified anomalies is not utilized fully and erroneous information may also be introduced into the new dataset. To address this, several soft versions of above are then established. For example, before training the detection model, LBS-SVDD~\cite{Bo2014An} first evaluates the abnormal probability of each sample based on the proportion of anomalies in its neighbors, whereas ADOA~\cite{Zhang2018Anomaly} assigns likelihood values to each sample according to its own isolation and its similarity to identified anomalies. They enhance the use of known information and simultaneously reduce the introduction of error messages. However, the calculation of probability may require high computational costs on large datasets and tends to be affected by the “curse of dimensionality” on high-dimensional datasets. Therefore, we propose a GAN-based model for semi-supervised outlier detection with few identified anomalies, which can utilize the potential information in identified anomalies directly.

\subsection{GAN-Based Outlier Detection Methods}
\label{sec:GAN}

GAN~\cite{Goodfellow2014Generative} is an adversarial representation learning model that has achieved state-of-the-art performance in various applications. For unsupervised outlier detection and semi-supervised outlier detection with only normal examples, GAN-based reconstruction model and generation model have been studied. GAN-based reconstruction models usually learn the generation mechanism of normal data by training a regular GAN~\cite{Schlegl2017Unsupervised} or a combination of GAN and autoencoder~\cite{Akcay2018GANomaly,Zenati2018Efficient,Adversarially2018Sabokrou}, and then measure the abnormal degree of example based on the reconstruction loss or discriminator loss. Moreover, in order to prevent slight anomalies from being reconstructed, Bian \etal ~\cite{Bian2019A} also perform active negative training to limit network generative capability. GAN-based generation models usually use the GAN to generate informative potential outliers~\cite{Liu2019Generative,Wang2018Anomaly} or infrequent normal samples~\cite{Swee2018DOPING}, such that subsequent detectors can describe a correct boundary. For supervised outlier detection, GAN~\cite{Fiore2017Using,Lima2019Heartbeat} is often used to synthesize minority class examples to balance the relative proportion between the two classes. Besides, Zheng \etal ~\cite{Zheng2018Generative} also take advantage of an adversarial deep denoising autoencoder to better extract latent representation of labeled transactions, which can greatly improve the accuracy of fraud detection. However, so far, there are few GAN-based studies focusing on semi-supervised outlier detection with few identified anomalies. Although Kimura \etal ~\cite{Kimura2018Semi} utilize both noisy normal images and given abnormal images for visual inspection, its main purpose is to eliminate the impact of abnormal pixels in some normal images during the reconstruction process, which differs considerably from our model.

\section{Methodology}
\label{sec:Methodology}
In this section, we first introduce the detection principle of the Artificially Generating Potential Outliers (AGPO)-based unsupervised outlier detection method MO-GAN~\cite{Liu2019Generative}, which are necessary to comprehend our proposed methods. Then two semi-supervised outlier detection models (\ie Dual-GAN and RCC-Dual-GAN) are proposed to effectively improve the detection rate of outliers.

\subsection{Background on MO-GAN}
\label{sec:Background}

\begin{figure}[ht]
	\centering
	\subfigure[\scriptsize{AGPO on
	Low.}]{
    	\label{fig:f1_1}
		\includegraphics[width=0.136\textwidth]{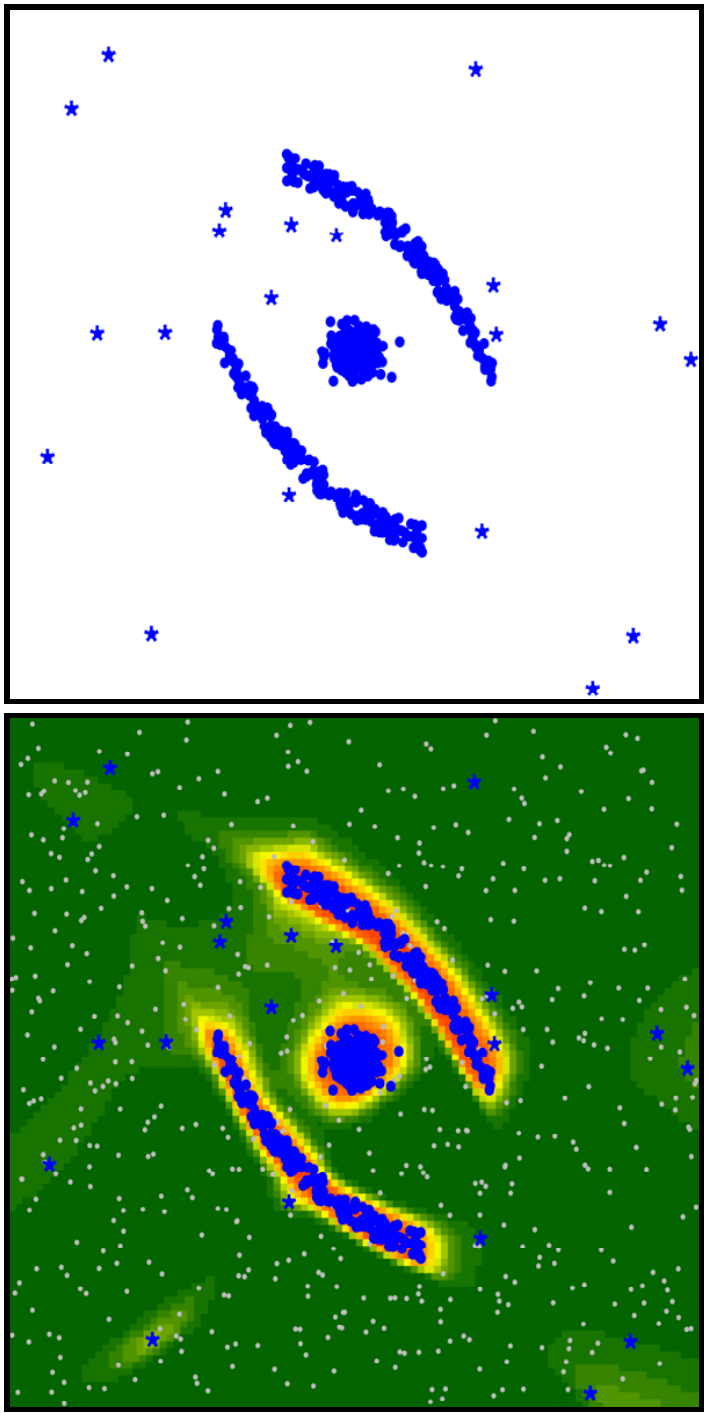}}
	\subfigure[\scriptsize{AGPO on High.}]{
    	\label{fig:f1_2}
		\includegraphics[width=0.136\textwidth]{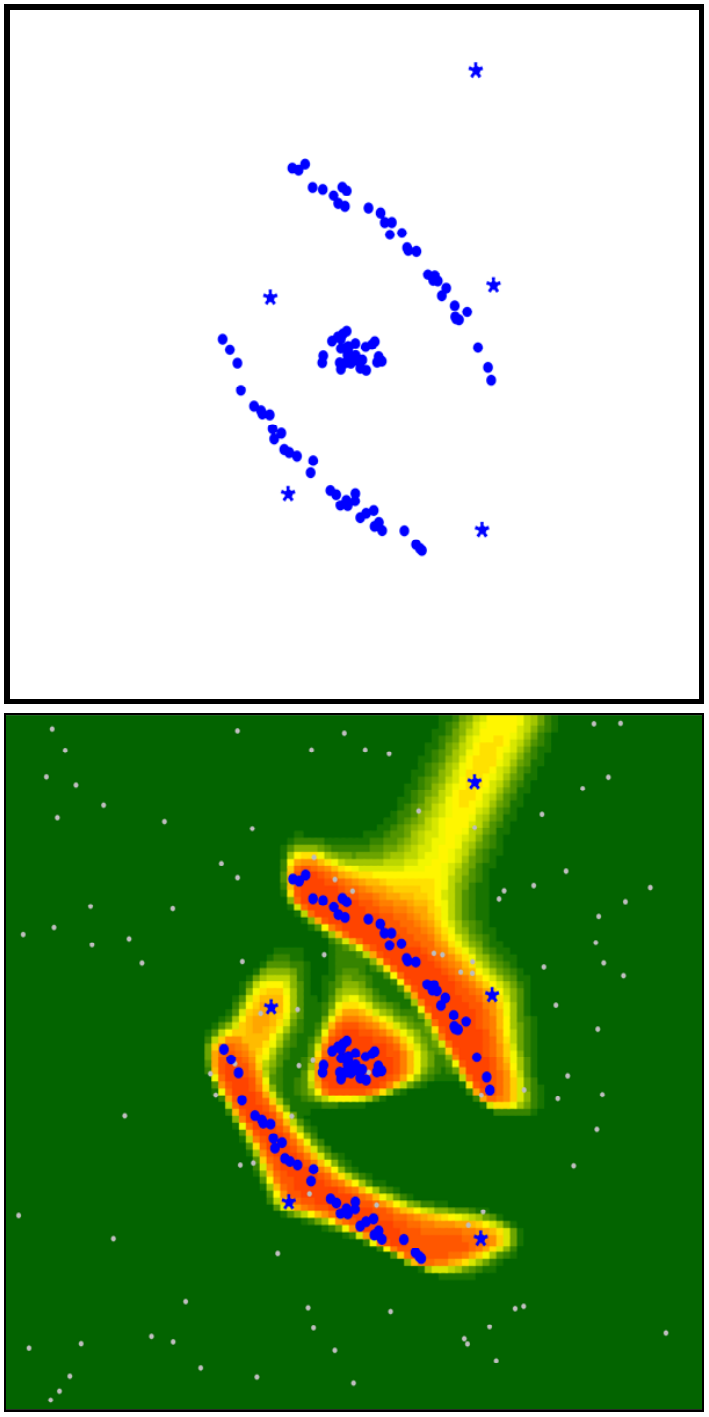}}
	\subfigure[\scriptsize{MO-GAN on High.}]{
    	\label{fig:f1_3}
		\includegraphics[width=0.136\textwidth]{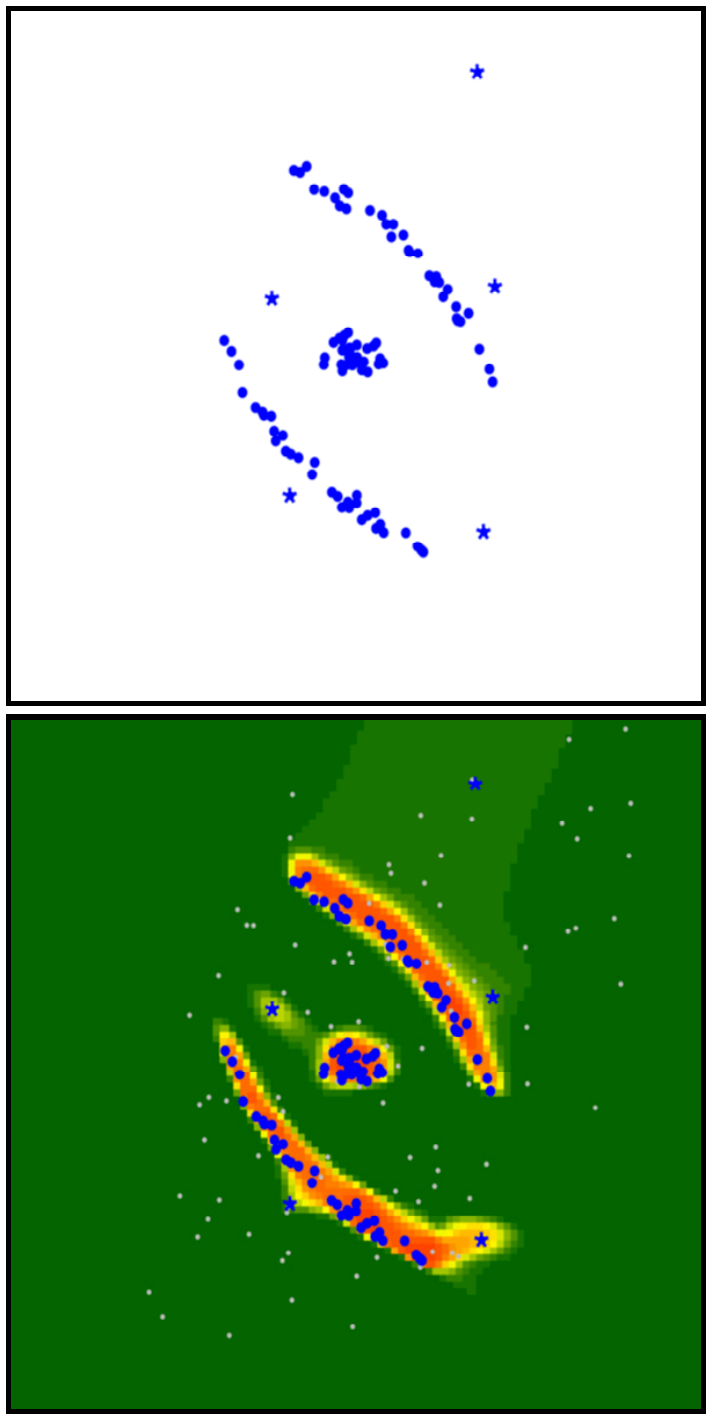}}
    \vspace{-10pt}
	\caption{Illustration of the detection performance of AGPO and MO-GAN. Normal points, outliers, and potential outliers are shown with blue dots, blue stars, and gray dots, respectively. High-dimensional data are presented as cross-sectional data, and data points closer to the green area are more likely to be outliers.}
    \vspace{-5pt}
	\label{fig:f1}
\end{figure}

Unsupervised outlier detection can be regarded as a density level detection process due to its default assumption. Unlike existing model- or proximity-based outlier detection, AGPO-based algorithms approach density level detection as a classification problem. First, numerous data points are randomly sampled as the potential outliers $x'$ (shown with gray dots in Fig. \ref{fig:f1}) to construct a reference distribution $\mu$. Then a classifier $\mathcal{C}$ is trained on the new dataset to separate potential outliers $x'_i$ from the original data $x_i\in X$ (shown with blue dots and stars in Fig. \ref{fig:f1}). In order to minimize the loss function $\mathcal{L}_\mathcal{C}$,
\begin{equation}
\begin{aligned} 
	\mathcal{L}_\mathcal{C}=-\sum_{i=1}^n[\log(\mathcal{C}(x_i))+\log(1-\mathcal{C}(x'_i))]
\end{aligned}
\end{equation}
the classifier $\mathcal{C}$ should assign a higher value to the original data $x$ having a higher relative density $\frac{\rho(x)}{\rho(x')}$, and a lower value to the opposite case. Thus, when faced with the uniform reference distribution $\mu$, the classifier $\mathcal{C}$ can describe a division boundary that encloses the concentrated normal samples $\{x_i | y_i=1\}$ (as shown in Fig. \ref{fig:f1_1}). 

However, when the dimension increases, a limited number of potential outliers $x'$ ($\rho(x')\Rightarrow 0$) cannot provide sufficient information for the classifier $\mathcal{C}$ to describe a correct boundary (as shown in Fig. \ref{fig:f1_2}). Therefore, MO-GAN was proposed to generate informative potential outliers directly to construct a reasonable reference distribution, which can ensure the relative density level of the normal case is greater than that of the outlier.
\begin{figure}[ht]
	\centering
	\vspace{-10pt}
	\includegraphics[scale=0.265]{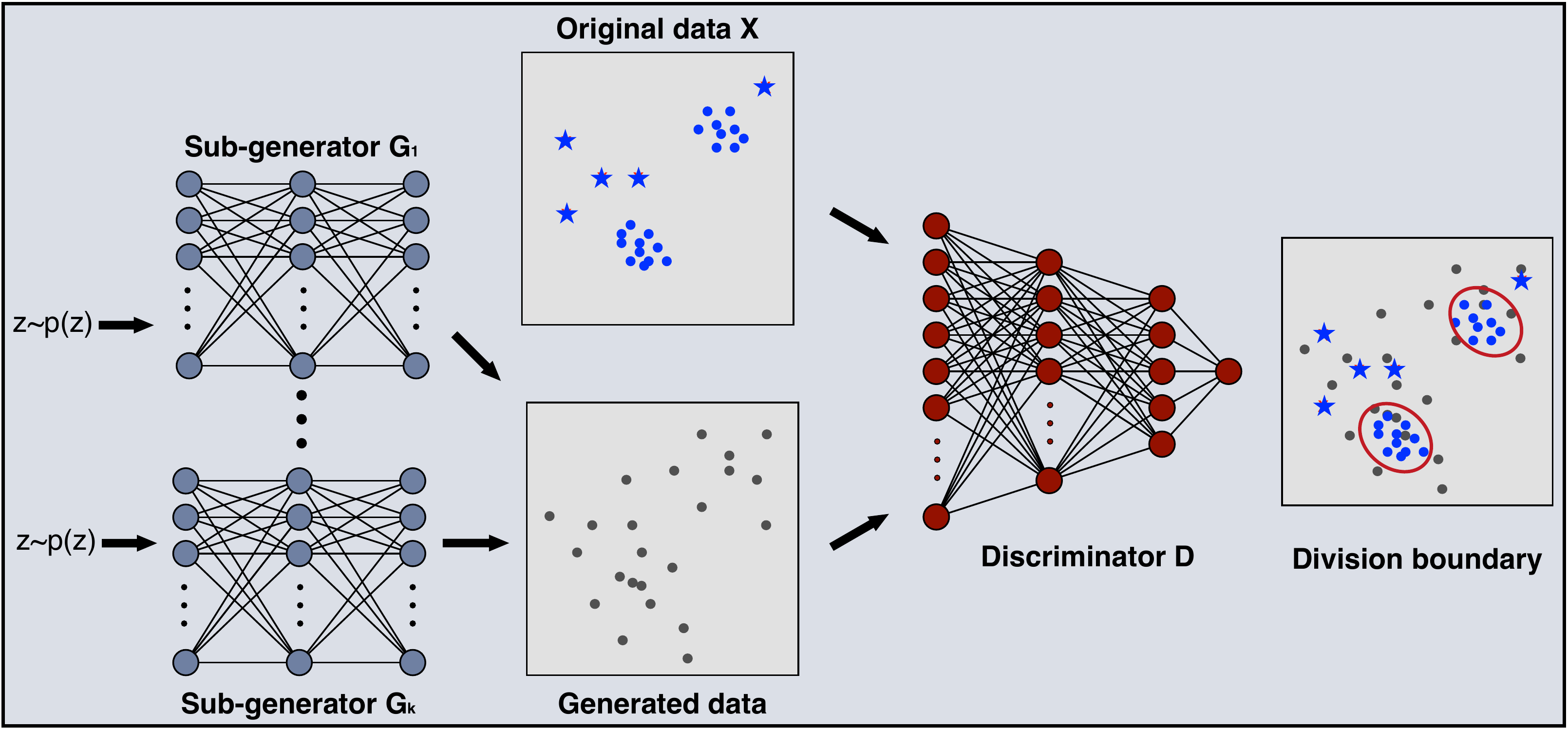}
	\vspace{-10pt}
	\caption{Illustration of the model details of MO-GAN. Through a dynamic game between $G_{1:k}$ and $D$, the specific sub-generator $G_j$ generates informative potential outliers that occur inside or close to the specific subsets $X_j$, and the discriminator $D$ describes a division boundary to enclose the concentrated original data.}
	\vspace{-5pt}
	\label{fig:f2}
\end{figure}

MO-GAN (shown in Fig. \ref{fig:f2}) consists of $k$ sub-generators $G_{1:k}$ and a
discriminator $D$. Its central idea is to let the specific sub-generator $G_j$ actively learn the generation mechanism of the data $x$ in the specific subset $X_j$, and gradually generate potential outliers $G_j(z)$ that occur inside or close to the data $x \in X_j$. Thus, the integration of different numbers $n_j$ of potential outliers $G_j(z)$ can provide a reasonable reference distribution $\mu$ for the whole dataset. More specifically, due to samples with similar outputs $D(x)$ are more likely to be similar, MO-GAN first divides the original dataset $X$ equally into $k$ subsets $X_{1:k}$ based on their similar outputs. Then, a dynamic game is executed between the sub-generators $G_{1:k}$ and discriminator $D$. Each sub-generator $G_j$ attempts to learn the generation mechanism of $X_j$ by making the generated samples $G_j(z)$ output similar values to $x\in X_j$, whereas discriminator $D$ attempts to identify the generated outliers $G_j(z)$ from the original data $x$, such as the classifier $\mathcal{C}$ in AGPO. Eventually, the MO-GAN reaches a Nash equilibrium through several iterations. Integrated different numbers of informative potential outliers $G_{1:k}(z)$ can construct a reasonable reference distribution $\mu$, and discriminator $D$ can describe a correct boundary to enclose concentrated original data (as shown in Fig. \ref{fig:f1_3} and Fig. \ref{fig:f2}).

\begin{figure*}[ht]
	\centering
	\includegraphics[scale=0.7]{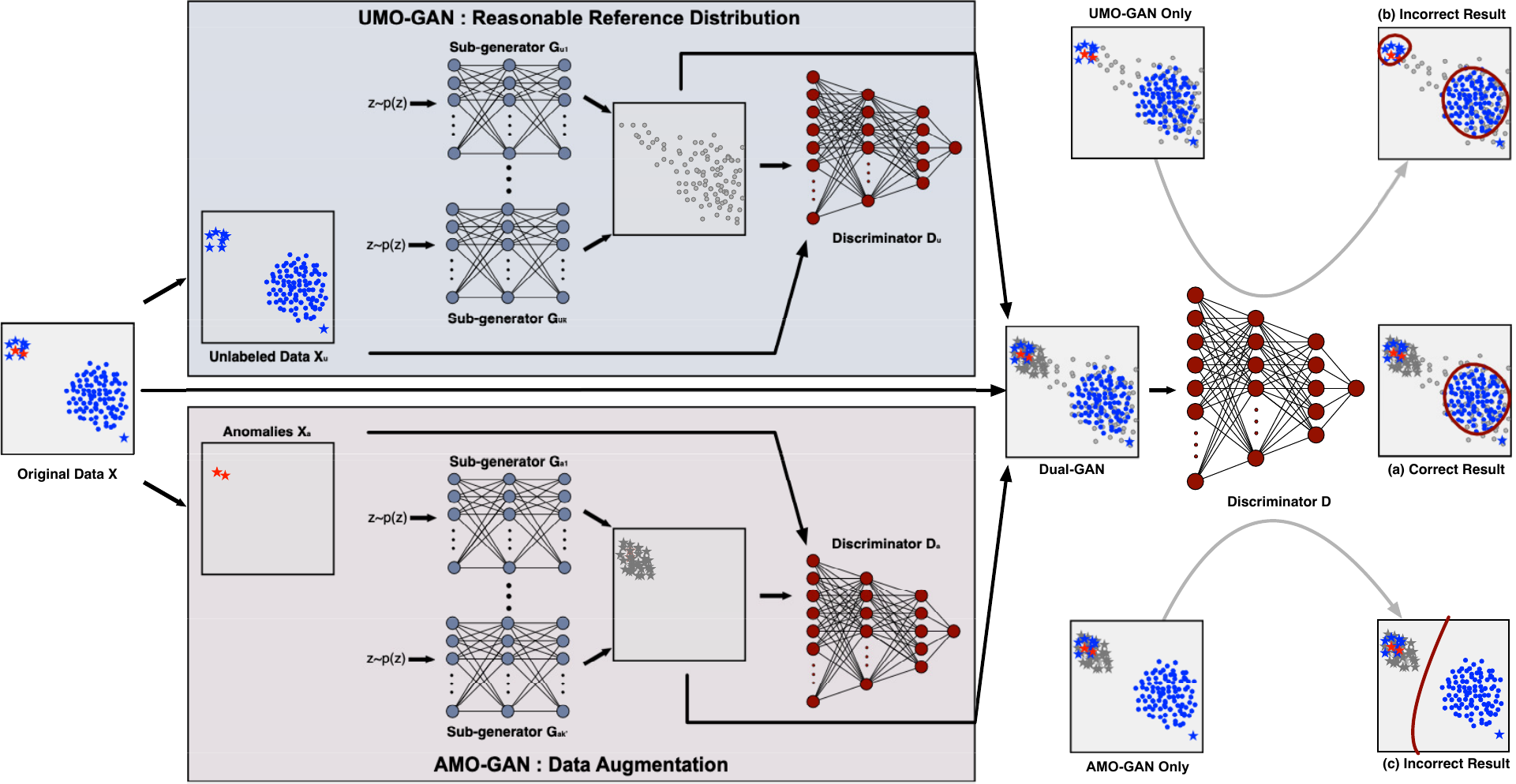}
	\vspace{-10pt}
	\caption{Illustration of model details of Dual-GAN. Normal examples, unidentified anomalies, and identified anomalies are shown with blue dots, blue stars, and red stars, respectively. The UMO-GAN is used to construct a reasonable reference distribution for unlabeled data, whereas the AMO-GAN is used to enhance the minority class. Thus, the overall discriminator will not only describe a division boundary that encloses the concentrated data, but also separate the partially identified group anomalies from the concentrated data.}
	\vspace{-15pt}
	\label{fig:f3}
\end{figure*}

\subsection{Outlier Detection with Few Identified Anomalies}
\label{sec:Semi}

The largest problem with unsupervised outlier detection (including MO-GAN) is that it cannot detect group anomalies in the absence of additional information. All labels and sufficient anomalies are difficult to obtain, but few common anomalous behaviors (\eg DoS and DDoS attacks) that have triggered alarms can be collected easily in many applications. These identified anomalies not only contain their own labels, but also potentially provide a priori information for other samples that with the same generation mechanism as identified anomalies. If these information is utilized fully, partially identified group anomalies will be detected accurately along with the discrete anomalies. Therefore, this section proposes two semi-supervised outlier detection approaches, namely, Dual-GAN and RCC-Dual-GAN, which can improve the detection accuracy by directly utilizing the potential information in identified anomalies.

\subsubsection{Dual-GAN}
\label{sec:Dual}

Assume a dataset $X=\{x_1, \dots, x_l, x_{l+1}, \dots, x_n\}$ with $l$ identified anomalies $X_a=\{x_i|y_i=0\}$ and $n-l$ unlabeled samples $X_u=\{x_i|y_i=1\ or\ 0\}$, where $x \in \mathbb{R}^d$ represents a data point, $y_i\in\{0,1\}$ represents its label, and $l \ll n$. Our goal is to identify a scoring function $\zeta(x)\in [0,1]$ that can assign a higher value (close to 1) to normal data and a lower value (close to 0) to the outlier. Because of a few identified anomalies, this scoring function should satisfy two conditions: (i) Based on the default assumption that outliers are not concentrated, the scoring function should output higher values to samples with higher density levels and output lower values to discrete data. (ii) Assuming that samples with the same generation mechanisms as identified anomalies are more likely to be outliers, the scoring function should output a value close to 0 for them and the identified anomalies. Thus, we first propose the Dual-GAN (shown in Fig. \ref{fig:f3}), which consists of two MO-GAN (\ie UMO-GAN and AMO-GAN) and an overall discriminator $D$.

The UMO-GAN attempts to generate samples that occur inside or around the target data to construct a reasonable reference distribution for unlabeled data. It takes unlabeled samples as input and includes $k$ sub-generators $G_{u1:uk}$ and a discriminator $D_u$. The specific sub-generator $G_{uj}$ learns the generation mechanism of the data $x\in X_{uj}$ by making the generated samples $G_{uj}(z)$ output similar values to $x\in X_{uj}$, whereas the discriminator $D_u$ guides the learning of sub-generator by identifying the generated samples $G_{u1:uk}(z)$ from unlabeled data $X_u$. The optimization framework of UMO-GAN is formulated as follows:
\begin{equation}
\begin{aligned}
	\min_{\theta_{d_u}}V_{D_u}=
	&-[\sum_{x\in X_u}\log(D_u(x))\\
	&+\sum_{j=1}^k\sum_{i=1}^{\left \lceil (n-l)/k \right \rceil}\log(1-D_u(G_{uj}(z_j^{(i)})))]
\end{aligned}
\end{equation}
\begin{equation}
\begin{aligned}
	\min_{\theta_{g_{uj}}}V_{G_{uj}}=
	&-\sum_{i=1}^{n-l}[T_{uj}\log(D_u(G_{uj}(z_j^{(i)})))\\
	&+(1-T_{uj})\log(1-D_u(G_{uj}(z_j^{(i)})))]
\end{aligned}
\end{equation}
where $T_{uj}$ is a representative statistic of $D_u(x|x\in X_{uj})$ (\eg the minimum value). With the iteration between $G_{uj}$ and $D_u$, the sub-generator $G_{uj}$ gradually generates informative potential outliers. And ultimately, when the dynamic game reaches the Nash equilibrium, the integration of different numbers $n_{uj}$ of potential outliers $G_{u1:uk}(z)$ (shown with gray dots in Fig. \ref{fig:f3}) provides a reasonable reference distribution for the unlabeled dataset $X_u$.

The AMO-GAN is used to generate samples similar to the identified anomalies to prevent the overall discriminator from overfitting or forgetting when dealing with the minority class~\cite{Gao2018Low}. It takes $l$ identified anomalies as input and includes $k'=\min(k,l)$ sub-generators $G_{a1:ak'}$ and a discriminator $D_a$. Specific sub-generator $G_{aj}$ learns the generation mechanism of the data $x\in X_{aj}$, and discriminator $D_a$ identifies the generated samples $G_{a1:ak'}(z)$ from identified anomalies $X_a$. The optimization framework of AMO-GAN is formulated as follows:
\begin{equation}
\begin{aligned}
	\min_{\theta_{d_a}}V_{D_a}=
	&-[\sum_{x\in X_a}\log(D_a(x))\\
	&+\sum_{j=1}^{k'}\sum_{i=1}^{\left \lceil l/k' \right \rceil}\log(1-D_a(G_{aj}(z_j^{(i)})))]
\end{aligned}
\end{equation}
\begin{equation}
\begin{aligned}
	\min_{\theta_{g_{aj}}}V_{G_{aj}}=
	&-\sum_{i=1}^{l}[T_{aj}\log(D_a(G_{aj}(z_j^{(i)})))\\
	&+(1-T_{aj})\log(1-D_a(G_{aj}(z_j^{(i)})))]
\end{aligned}
\end{equation}
where $T_{aj}$ is a representative statistic of $D_a(x|x\in X_{aj})$. Unlike UMO-GAN, it will continue training after it reaches the Nash equilibrium because the purpose of the AMO-GAN is to generate data points as similar as possible to the identified anomalies. Finally, the integration of numerous $n_{aj}$ of potential outliers $G_{a1:ak'}(z)$ (shown with gray stars in Fig. \ref{fig:f3}) can augment the minority class to ensure that partially identified group anomalies are detected as anomalies.

The overall discriminator $D$, which takes all original data and generated potential outliers as input, attempts to describe an accurate division boundary by identifying all potential outliers (\ie $G_{u1:uk}(z)$ and $G_{a1:ak'}(z)$) and identified anomalies $X_a$ from the unlabeled data $X_u$. The optimization function of $D$ is formulated as follows:
\begin{equation}
\begin{aligned}
	\min_{\theta_d}V_D=
	&-[\sum_{x\in X_u}\log(D(x))+\sum_{x\in X_a}\log(1-D(x))\\
	&+\sum_{j=1}^k\sum_{i=1}^{n_{uj}}\log(1-D(G_{uj}(z_j^{(i)})))\\
	&+\sum_{j=1}^{k'}\sum_{i=1}^{n_{aj}}\log(1-D(G_{aj}(z_j^{(i)})))]
\end{aligned}
\end{equation}
where $n_{uj}$ and $n_{aj}$ represent the number of potential outliers generated by $G_{uj}$ and $G_{aj}$, respectively. More potential outliers $G_{uj}(z)$ must be generated for the less concentrated subset $X_{uj}$ to create a reasonable reference distribution~\cite{Liu2019Generative}. At the beginning of the iteration, randomly generated potential outliers may not provide sufficient information for $D$. However, when the two MO-GAN models reach the Nash equilibrium, the integration of different numbers $n_{uj}$ of potential outliers $G_{u1:uk}(z)$ can provide a reasonable reference distribution for the unlabeled data $X_u$, whereas the integration of numerous $n_{aj}$ of potential outliers $G_{a1:ak'}(z)$ can augment the minority class. Thus, in order to minimize the optimization function $V_D$, the overall discriminator $D$ will not only assign a higher value (close to 1) to concentrated unlabeled data, but also assign a lower value (close to 0) to discrete anomalies and partially identified group anomalies (shown in Fig. \ref{fig:f3}(a)), which is the scoring function $\zeta(x)$ we are looking for. Compared with unsupervised detection using only UMO-GAN (shown in Fig. \ref{fig:f3}(b)), Dual-GAN can also detect group anomalies with the help of few identified anomalies. Compared with supervised detection using only AMO-GAN (shown in Fig. \ref{fig:f3}(c)), Dual-GAN can also detect previously unknown discrete anomalies.

In addition, two issues that have a substantial effect on the results, namely, the evaluation of Nash equilibrium and the selection of optimal model, must be discussed to ensure a more intelligent and reliable detection.

\begin{figure*}[ht]
	\centering
	\includegraphics[scale=0.7]{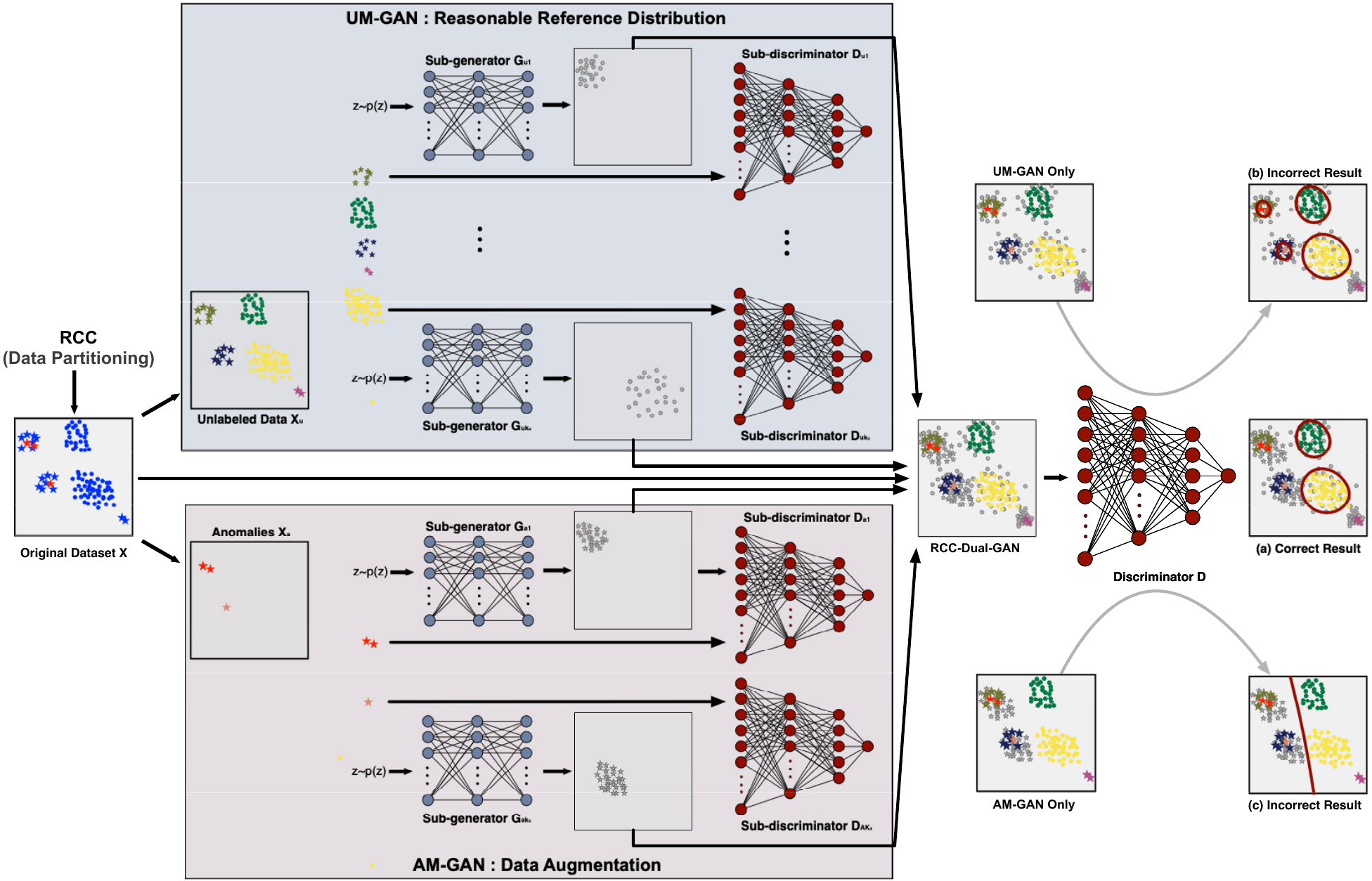}
	\vspace{-10pt}
	\caption{Illustration of the model details of RCC-Dual-GAN. RCC first divides the unlabeled data and identified anomalies into different subsets, which are represented by different colored dots and stars. Thus, UM-GAN and AM-GAN can directly learn the generation mechanism of different subsets.}
	\vspace{-15pt}
	\label{fig:f4}
\end{figure*}

\textbf{Nash} \textbf{Equilibrium} in GAN means that the distribution of the real data has been learned by the generator, and the discriminator cannot recognize the difference between the two distributions. The original GAN uses the classification error to evaluate the similarity between the generated data and the real data, that is, the Nash equilibrium is reached when the error is close to $-\frac{1}{2}\log(\frac{1}{2})$. However, the absolute Nash equilibrium cannot be guaranteed when the objective function is non-convex. The previously proposed MO-GAN utilizes the trend of the generator loss to evaluate their similarity, that is, the Nash equilibrium is reached when the downward trend of generator loss tends to be slow. However, accurate assessment of the trend requires human intervention due to the fluctuation of the loss. Therefore, we propose an evaluation indicator, Nearest Neighbor Ratio ($NNR$), to directly measure the similarity between the two distributions. First, $p$ samples $x_i$ are selected randomly from one subset, and the ratio $nnr_i$ of data $x'$ belonging to another subset among the $q$ nearest neighbors of each sample $x_i$ is calculated. If $nnr_i$ is greater than a certain threshold $\tau_1$, the sample $x_i$ can be thought of as having a similar generation mechanism to the data $x'$ in another subset. Then calculate the ratio $NNR$ of the samples $x_i$ that have a similar generation mechanism to $x'$ in the randomly selected $p$ samples. If the $NNR$ is greater than $\tau_2$, the two subsets are considered to be generated from similar distributions and the dynamic game reaches the Nash equilibrium.

\textbf{Optimal} \textbf{model} refers to the model that can most effectively identify outliers from the whole dataset during iteration. Given no additional information, the evaluation of detection performance and the selection of the optimal model are difficult for unsupervised outlier detection. Fortunately, the data used to train the semi-supervised outlier detection model usually contain few identified anomalies, which can provide valuable guidance for the selection of the final model. In this paper, we use the Average Position ($AP$) of known anomalies in the ascending order of all real data output results $D(x)$ to measure the performance of the overall discriminator $D$. A lower $AP$ means that the model assigns lower values to identified anomalies than to others, and the model corresponding to the lowest $AP$ is used as the final model for subsequent detection.

\subsubsection{RCC-Dual-GAN}
\label{sec:RCC}

In general, Dual-GAN can achieve good detection performance. However, as the cluster structure of the data becomes more complex, instances with similar output values may not all be similar to one another in the sample space, that is, the data points divided according to their similar outputs are not necessarily close to each other, and the generated data whose outputs are similar to that of target data are not necessarily similar to the target data. Therefore, we then propose a modified model RCC-Dual-GAN based on Dual-GAN to create the reference distribution and augment the minority class more robustly. The network structure and detection process of RCC-Dual-GAN are illustrated in Fig. \ref{fig:f4}, where the unlabeled data $X_u$ and identified anomalies $X_a$ are first divided into different subsets by RCC.

\textbf{RCC}~\cite{Shah2017Robust} is a non-parametric clustering that can achieve high clustering accuracy across multiple domains without knowing the number of clusters. Given the unlabeled data $X_u$ as an example, RCC first constructs a connectivity structure $\mathcal{E}_u$ based on mutual $k$-nearest neighbor connectivity. And then, a set of representatives $U_u=\{u_1,\dots,u_{n-l}\}$ of the unlabeled data $X_u=\{x_1,\dots,x_{n-l}\}$ is optimized to reveal the cluster structure latent in $X_u$. The representative $u_i$ should be as similar as possible to the corresponding unlabeled data $x_i$, and the representatives $(u_p,u_q)$ of interconnected data $(x_p,x_q)\in \mathcal{E}_u$ should be as similar as possible. The optimization objective is formulated as follows:
\begin{equation}
\begin{aligned}
	\min_uC(U_u)=&\frac{1}{2}\sum_{i=1}^{n-l}\left \| x_i-u_i \right \|_2^2\\
	&+\frac{\lambda}{2}\sum_{(x_p,x_q)\in \mathcal{E}_u}\omega_{p,q}\rho(\|u_p-u_q\|_2)
\end{aligned}
\end{equation}
where $\lambda$ is used to balance the strength of different objective terms, $\omega_{p,q}$ is used to balance the contribution of each point to the pairwise terms, and $\rho(\cdot)$ is a penalty on the regularization terms. Finally, based on the optimized $U_u$, RCC constructs a graph $\mathcal{G}_u$ in which a pair $x_p$ and $x_q$ is connected if $\|u_p-u_q\|_2<\delta$, such that $k_u$ different unlabeled subsets $X_{u1:uk_u}$ are output. Compared with the subsets divided by similar outputs, the subsets partitioned by RCC can accurately reflect the cluster structure latent in the data even in the case of complex data structures. 

After the unlabeled data and identified anomalies are divided into $k_u$ and $k_a$ subsets, respectively, RCC-Dual-GAN replaces MO-GAN with M-GAN to create the reference distribution and augment the minority class in more detail. The UM-GAN includes $k_u$ sub-generators $G_{u1:uk_u}$ and sub-discriminators $D_{u1:uk_u}$. Each specific sub-GAN can directly learn the generation mechanism of the data $x\in X_{uj}$ through the dynamic game between $G_{uj}$ and $D_{uj}$,
\begin{equation}
\begin{aligned}
	\min_{\theta_{g_{uj}}}\max_{\theta_{d_{uj}}}V(D_{uj},G_{uj})=
	&\sum_{x\in X_{uj}}\log(D_{uj}(x))+\\
	&\sum_{i=1}^{n_{uj}}\log(1-D_{uj}(G_{uj}(z_j^{(i)})))
\end{aligned}
\end{equation}
where $n_{uj}$ represents the number of samples in the $j$th unlabeled subset. The AM-GAN includes $k_a$ sub-generators $G_{a1:ak_a}$ and sub-discriminators $D_{a1:ak_a}$. Each specific sub-GAN directly learns the deep representation of data $x\in X_{aj}$ through the dynamic game between $G_{aj}$ and $D_{aj}$,
\begin{equation}
\begin{aligned}
	\min_{\theta_{g_{aj}}}\max_{\theta_{d_{aj}}}V(D_{aj},G_{aj})=
	&\sum_{x\in X_{aj}}\log(D_{aj}(x))+\\
	&\sum_{i=1}^{n_{aj}}\log(1-D_{aj}(G_{aj}(z_j^{(i)})))
\end{aligned}
\end{equation}
where $n_{aj}$ represents the number of samples in $X_{aj}$. The overall discriminator $D$ still attempts to identify all potential outliers and identified anomalies from the unlabeled data,
\begin{equation}
\begin{aligned}
	\min_{\theta_d}V_D=
	&-[\sum_{x\in X_u}\log(D(x))+\sum_{x\in X_a}\log(1-D(x))\\
	&+\sum_{j=1}^{k_u}\sum_{i=1}^{n_{uj}'}\log(1-D(G_{uj}(z_j^{(i)})))\\
	&+\sum_{j=1}^{k_a}\sum_{i=1}^{n_{aj}'}\log(1-D(G_{aj}(z_j^{(i)})))]
\end{aligned}
\end{equation}
where $n_{uj}'$ and $n_{aj}'$ represent the number of potential outliers generated by $G_{uj}$ and $G_{aj}$, respectively. The UM-GAN will generate the same number of potential outliers for different unlabeled subsets, which is different from the UMO-GAN. Because each unlabeled data subset partitioned by RCC contains a different number of samples, and the concentrated data are usually divided into large subsets.

At the beginning of the iteration, the two M-GANs randomly generate potential outliers in the sample space, whereas the overall discriminator $D$ describes a rough boundary to separate them from unlabeled data. However, when all sub-GANs reach the Nash equilibrium, the integration of the same number of potential outliers (shown with gray dots in Fig. \ref{fig:f4}) generated by $G_{u1:uk_u}$ can provide a reasonable reference distribution for the unlabeled data, and the integration of numerous potential outliers (shown with gray stars in Fig. \ref{fig:f4}) generated by $G_{a1:ak_a}$ can augment the minority class. Consequently, the overall discriminator $D$ will not only describe a division boundary that encloses the concentrated data but also separate the partially identified group anomalies from the concentrated data (shown with the red lines in Fig. \ref{fig:f4}(a)). Compared with the potential outliers generated by outputting similar values, the potential outliers generated by directly learning can more effectively assist the overall discriminator $D$ in describing a correct boundary even in the case of complex data structures.

\renewcommand{\algorithmicrequire}{ \textbf{Input:}}
\renewcommand{\algorithmicensure}{ \textbf{Output:}}
\begin{algorithm}[htb] 
\caption{RCC-Dual-GAN} 
\label{alg:RCC-Dual-GAN} 
\begin{algorithmic}[1] 
\Require 
$X=\{X_a,X_u\}$; $p_z$; $I$; $\tau_1$; $\tau_2$; $AP(D'(X))$
\Ensure 
outlier score, $OS(x)$
\State \textbf{Divide} $X_u$ into $k_u$ subsets $X_{u1:uk_u}$
\State \textbf{Divide} $X_a$ into $k_a$ subsets $X_{a1:ak_a}$
\State \textbf{Initialize} $G_{u1:uk_u}$; $D_{u1:uk_u}$; $G_{a1:ak_a}$; $D_{a1:ak_a}$; $D$; $m_{uj}$; $m_{aj}$; $m_{uj}'$; $m_{aj}'$
\Repeat
\For {$j=1$ to $k_u$}
\State Sample $m_{uj}$ noises $z$ from $p_z$
\State Sample $m_{uj}$ samples $x$ from $X_{uj}$
\State Compute $NNR_j$ by $\tau_1$, $G_{uj}(z)$ and $x \in X_{uj}$
\If {$NNR_j< \tau_2$} 
\State Update $G_{uj}$ and $D_{uj}$ by optimizing Eq. (8)
\EndIf
\State Sample $m_{uj}'$ noises $z$ from $p_z$
\EndFor 
\For {$j=1$ to $k_a$}
\State Sample $m_{aj}$ noises $z$ from $p_z$
\State Sample $m_{aj}$ samples $x$ from $X_{aj}$
\State Update $G_{aj}$ and $D_{aj}$ by optimizing Eq. (9)
\State Sample $m_{aj}'$ noises $z$ from $p_z$
\EndFor 
\State Update $D$ by optimizing Eq. (10)
\If {$AP(D(X)) < AP(D'(X))$} 
\State Save $D$ as $D'$
\EndIf
\Until the maximum number of iterations $I$
\State $OS(x)=1-D'(x)$\\
\Return $OS(x)$ 
\end{algorithmic} 
\end{algorithm}

\section{Experiments and Applications}
\label{sec:4}
Extensive experiments are conducted on synthetic data and real-world data to investigate the importance of the effective use of identified anomalies. In addition, we apply the proposed models to two practical tasks (\ie credit card fraud detection and network intrusion detection) to study the performance of different algorithms in complex situations.

\subsection{Experiments}
\label{sec:4.1}
\subsubsection{Baselines and Parameter Settings}
\label{sec:4.1.1}
We compare the proposed models (\ie Dual-GAN and RCC-Dual-GAN) with several representative outlier detection algorithms. (i) Three of the most common unsupervised approaches ($k$NN, LOF, $k$-means) are first selected because their effectiveness and robustness have been proven in multiple performance evaluations. (ii) The basic model MO-GAN, which utilizes the explicit information and guidance information in identified anomalies, is performed to investigate the significance of the data augmentation in Dual-GAN. (iii) The supervised Sup-GAN~\cite{Fiore2017Using}, which uses GAN to increase the relative proportion of the minority class, is used to explore the importance of the unsupervised module in Dual-GAN. (iv) The extended supervised Sup-RCC-GAN, where the single GAN in Sup-GAN is replaced by our proposed combination of RCC and M-GAN, is compared to further demonstrate the performance advantages of multiple GAN. (v) The semi-supervised ADOA~\cite{Zhang2018Anomaly}, which attaches a weight to each instance, is used to evaluate the performance of our proposed semi-supervised models.

For non-GAN-based models, we attempt to find the optimal parameters in a range of values. For example, the parameters $k$ in $k$NN and LOF are searched from 2 to $\left \lceil \frac{n}{10} \right \rceil$, the $k$ in $k$-means is selected from 1 to $\left \lceil \frac{n}{100} \right \rceil$, and the $\beta$ in ADOA is adjusted from 0.1 to 0.9. For all GAN-based models, we adopt a unified network structure: (i) five sub-generators against one discriminator for MO-GAN, UMO-GAN, and AMO-GAN; (ii) a three-layer network ($d*d*d$) for generator and a four-layer network ($d*\min(n,1000)*10*1$) for discriminator; (iii) Orthogonal initializer for generator and Variance-Scaling for discriminator; (v) $\tau_1$, $\tau_2$ and $I$ are set to 0.5, 0.4 and 1000, respectively; and (vi) the final model in Sup-GAN is selected by the accuracy, and the $AP$ is for others.

\subsubsection{Experiments on Synthetic Data}
\label{sec:4.1.2}
\begin{figure}[ht]
	\centering
    \subfigure[\scriptsize{Synthetic Dataset}]{
    	\label{fig:f7_1}
\includegraphics[width=0.225\textwidth]{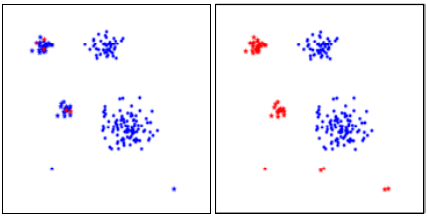}}
	\subfigure[\scriptsize{$k$-means}]{
    	\label{fig:f7_2}
\includegraphics[width=0.225\textwidth]{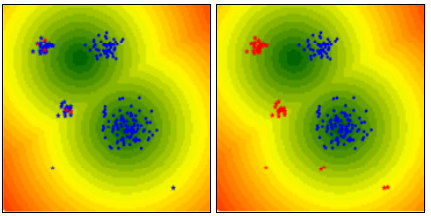}}
    \subfigure[\scriptsize{MO-GAN}]{
    	\label{fig:f7_3}
\includegraphics[width=0.225\textwidth]{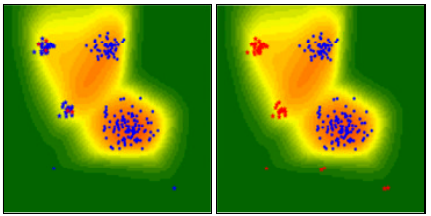}}
	\subfigure[\scriptsize{Dual-GAN}]{
    	\label{fig:f7_4}
\includegraphics[width=0.225\textwidth]{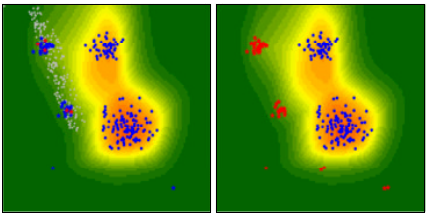}}
    \subfigure[\scriptsize{RCC-Dual-GAN}]{
    	\label{fig:f7_5}
\includegraphics[width=0.225\textwidth]{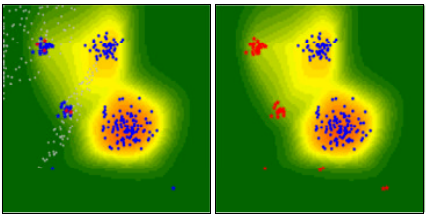}}
	\subfigure[\scriptsize{sup-GAN}]{
    	\label{fig:f7_6}
\includegraphics[width=0.225\textwidth]{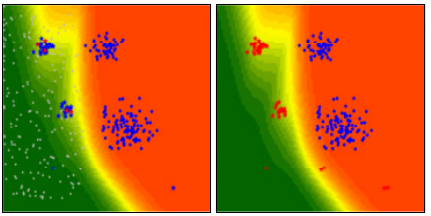}}
    \subfigure[\scriptsize{Sup-RCC-GAN}]{
    	\label{fig:f7_7}
\includegraphics[width=0.225\textwidth]{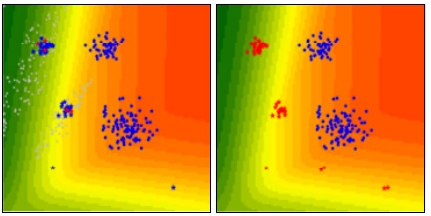}}
	\subfigure[\scriptsize{ADOA}]{
    	\label{fig:f7_8}
\includegraphics[width=0.225\textwidth]{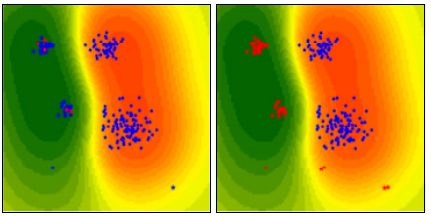}}
	\vspace{-5pt}
	\caption{Illustration of the synthetic datasets and detection results. The meaning of the color is the same as that of Fig. \ref{fig:f1}, and color comparisons only make sense in their own pictures.}
	\vspace{-10pt}
	\label{fig:f7}
\end{figure}

We generate a couple of datasets (\ie training dataset and test dataset) based on the usual assumptions of outliers to study the performance characteristics of different algorithms in more detail. The training dataset (as shown on the left in Fig. \ref{fig:f7_1}) consists of two sets of normal data, two sets of group anomalies, and two discrete anomalies. And, in order to match the setting of anomaly detection with few
identified anomalies, five examples are randomly sampled from all anomalies as the identified anomalies (shown with red stars). The test dataset (as shown on the right in Fig. \ref{fig:f7_1}) contains two sets of normal data, two sets of group anomalies, and five discrete anomalies. The normal data and group anomalies have exactly the same generation mechanisms with the training data, whereas the five discrete outliers are unidentified or emerging anomalies

\begin{figure}[ht]
	\centering
	\includegraphics[width=0.42\textwidth]{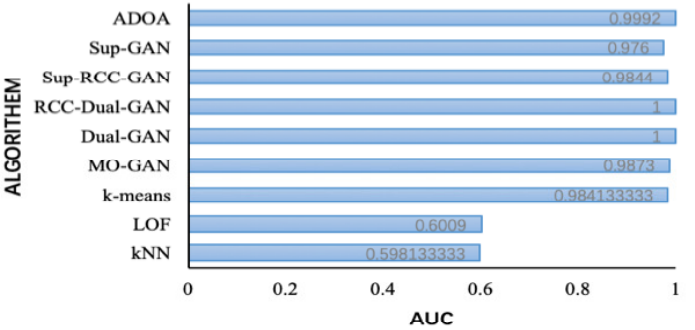}
	\vspace{-10pt}
	\caption{Experimental results on the synthetic dataset.}
    \vspace{-10pt}
    \label{fig:f6}
\end{figure}

The experimental results of our proposed methods and seven competitors are shown in Fig. \ref{fig:f6}. Dual-GAN and RCC-Dual-GAN obtain the best detection results (AUC=1), whereas $k$NN and LOF achieve very poor results because the two proximity-based methods with parameter $k$ in a specific range cannot identify group anomalies. As for the other five competitors, in order to clearly illustrate their performance characteristics, we provide a visual representation of the detection results as shown in Fig. \ref{fig:f7}. The cluster-based $k$-means (shown in Fig. \ref{fig:f7_2}) achieves the optimal result when $k=2$. However, the cluster centers of the two sets of normal data are not accurately identified due to the interference of unidentified anomalies. The basic model MO-GAN (shown in Fig. \ref{fig:f7_3}) describes a division boundary that encloses the concentrated data, such that the discrete anomalies can be accurately identified. However, partially identified group anomalies cannot be separated from the concentrated normal data because only explicit information in identified anomalies is used. Supervised Sup-GAN and Sup-RCC-GAN (shown in Fig. \ref{fig:f7_6} and \ref{fig:f7_7}) that use GAN to enhance the minority class can identify group anomalies represented by identified anomalies. However, the detection of discrete and emerging anomalies will face substantial challenges because the patterns of normal data are not established. The semi-supervised ADOA (shown in Fig. \ref{fig:f7_8}) that obtains the suboptimal AUC value can identify all anomalies in the training data, but the ADOA only divides the weighted normal data from the weighted anomalies, such that the detection results of emerging anomalies in the test data cannot be guaranteed. By contrast, our proposed models (shown in Fig. \ref{fig:f7_4} and \ref{fig:f7_5}) can describe a division boundary that encloses the normal data, showing evident advantages in identifying the partially identified group anomalies and all discrete anomalies. 

\subsubsection{Experiments on Real-world Data}
\label{sec:4.1.3}

Ten real-world datasets that often appear in other outlier detection literatures are selected for the following experiments to obtain an overall assessment of different algorithms. These datasets are first processed as outlier evaluation datasets according to the procedure described in~\cite{Campos2016On}. We then divide each dataset into a training dataset and a test dataset in the ratio of 2 to 1. Furthermore, 10\% of anomalies in the training data are randomly selected as identified anomalies to match the setting of few identified anomalies. Detailed information on these datasets is listed in Table \ref{tab:t_1}, where NoC. Indicates the number of identified anomalies clusters that are divided by RCC.
{
	\begin{table}[htb]
		\caption{Description of the Real-World Datasets}
		\vspace{-15pt}
		\label{tab:t_1}
		\centering
		\begin{tabular}[t]{p{1.2cm}p{0.87cm}p{0.5cm}p{0.5cm}p{0.5cm}p{0.9cm}p{1cm}}
        
			\hline
            \multirow{2}*{Dataset} & \multirow{2}*{Dim.}& \multicolumn{4}{l}{Training Date}& \multirow{2}*{Test Date}\\
            \cline{3-6}
             && \scriptsize{Nor.}& \scriptsize{Ano.} &\scriptsize{Ide.}& \scriptsize{NoC.(Ide.)}& \\
			\hline
            Thyroid & 6 & 2451 & 64 & 7 & 1 & 1255\\
            Pima & 8 & 328 & 179 & 18 & 3 & 261\\
            Stamps & 9 & 206 & 21 & 3 & 1 & 113\\
            Pageblocks & 10 & 3263 & 340 & 34 & 16 & 1790\\
            Cardio & 21 & 1103 & 118 & 12 & 3 & 610\\
            Waveform & 21 & 2229 & 67 & 7 & 1 & 2322\\
            Spambase & 57 & 1681 & 1120 & 112 & 12 & 1406\\
            Optdigits & 64 & 3377 & 100 & 10 & 1 & 2179\\
            Mnist  & 100 & 4602 & 467 & 47 & 14 & 2534\\
            Har & 561 & 1868 & 20 & 2 & 1 & 972\\
			\hline
		\end{tabular}
		\vspace{-5pt}
	\end{table}	
}

Experimental results on real-world datasets are shown in Table \ref{tab:t_2}. The highest AUC for each dataset is highlighted in bold. The average ranks of nine algorithms on ten datasets are provided in the last row of Table \ref{tab:t_2}.    

\begin{figure*}[ht]
	\centering
    \subfigure[\scriptsize{Thyroid}]{
    	\label{fig:f8_1}
\includegraphics[width=0.19\textwidth]{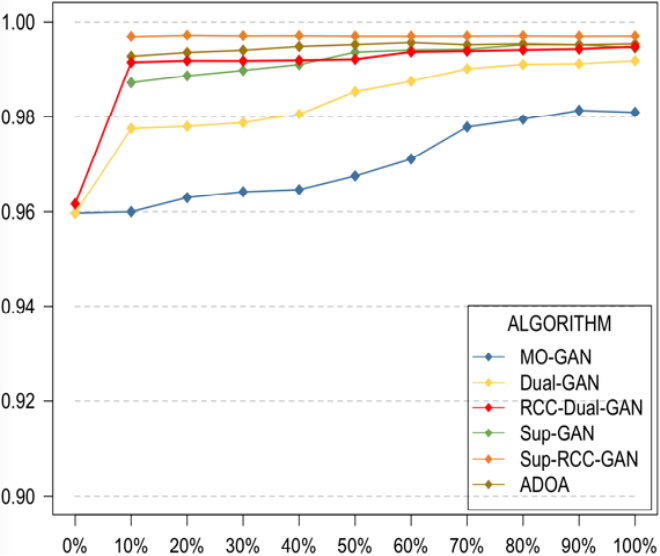}}
	\subfigure[\scriptsize{Pima}]{
    	\label{fig:f8_2}
\includegraphics[width=0.19\textwidth]{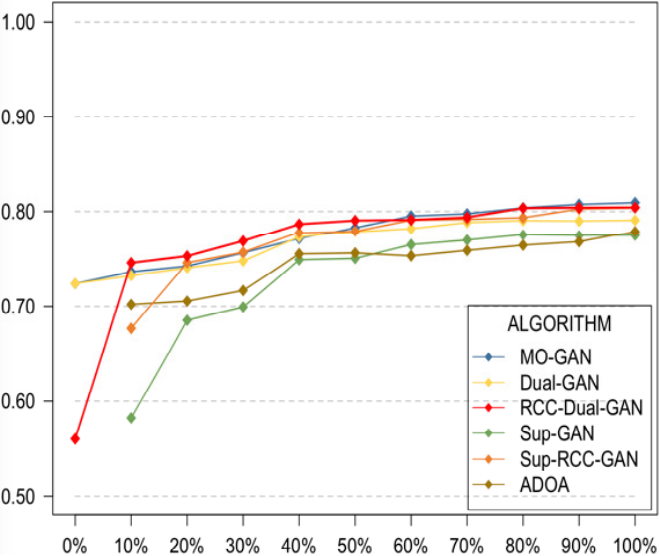}}
    \subfigure[\scriptsize{Stamps}]{
    	\label{fig:f8_3}
\includegraphics[width=0.19\textwidth]{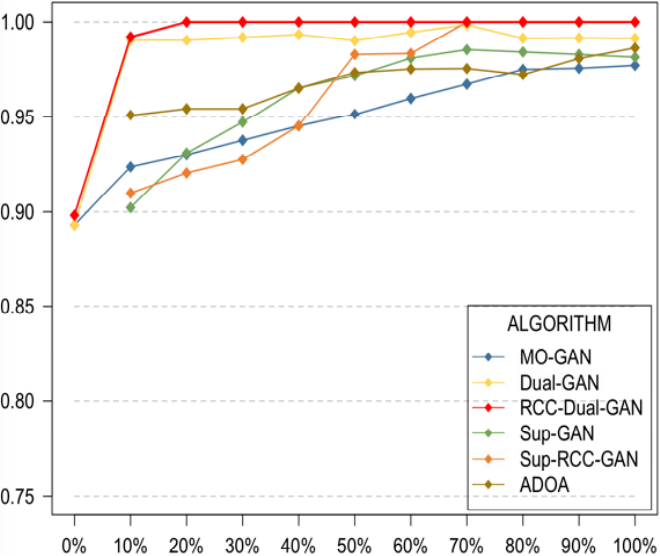}}
	\subfigure[\scriptsize{Pageblocks}]{
    	\label{fig:f8_4}
\includegraphics[width=0.19\textwidth]{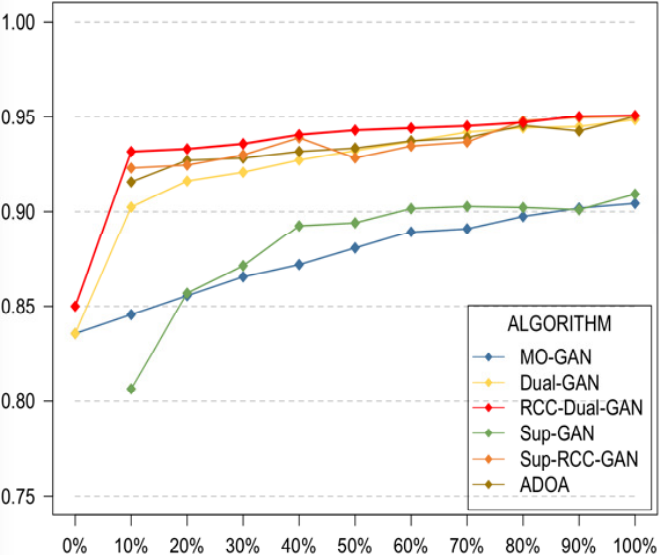}}
    \subfigure[\scriptsize{Cardio}]{
    	\label{fig:f8_5}
\includegraphics[width=0.19\textwidth]{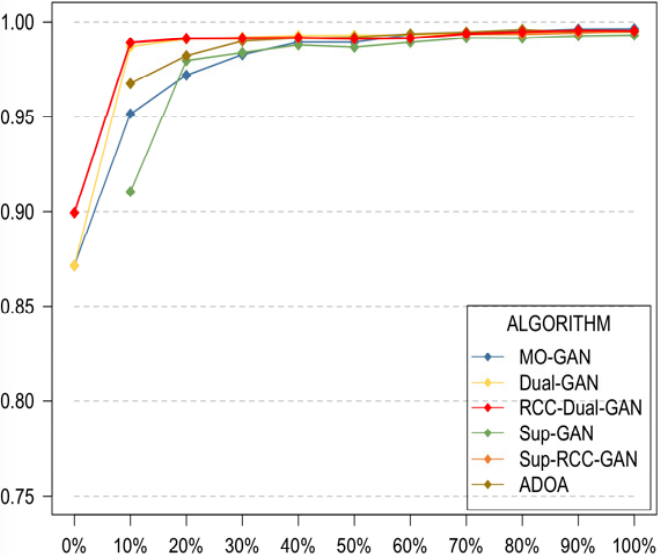}}
	\subfigure[\scriptsize{Waveform}]{
    	\label{fig:f8_6}
\includegraphics[width=0.19\textwidth]{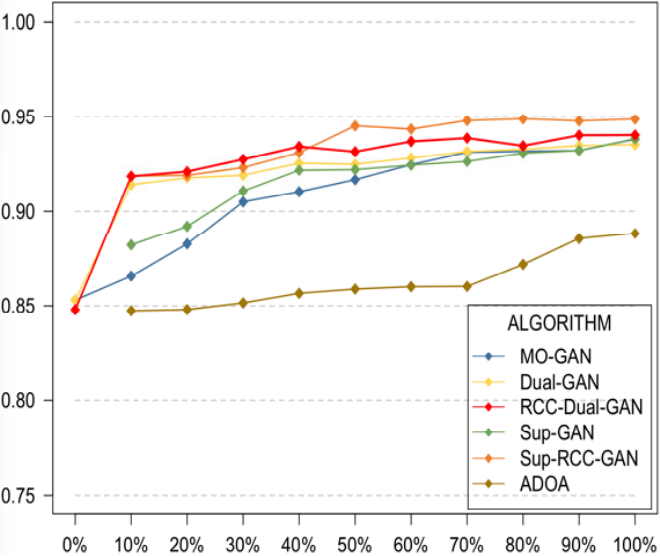}}
    \subfigure[\scriptsize{Spambase}]{
    	\label{fig:f8_7}
\includegraphics[width=0.19\textwidth]{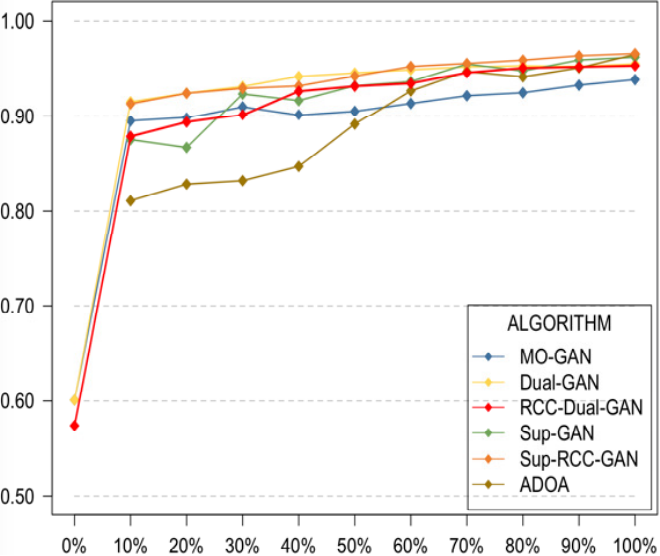}}
	\subfigure[\scriptsize{Optdigits}]{
    	\label{fig:f8_8}
\includegraphics[width=0.19\textwidth]{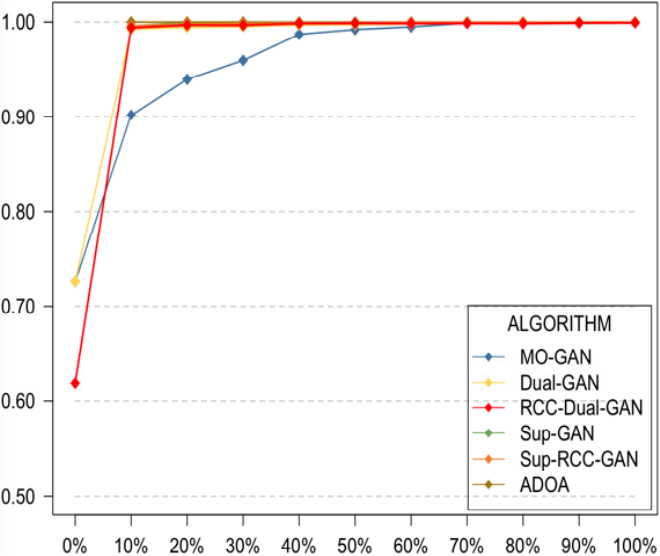}}
	\subfigure[\scriptsize{Mnist }]{
    	\label{fig:f8_9}
\includegraphics[width=0.19\textwidth]{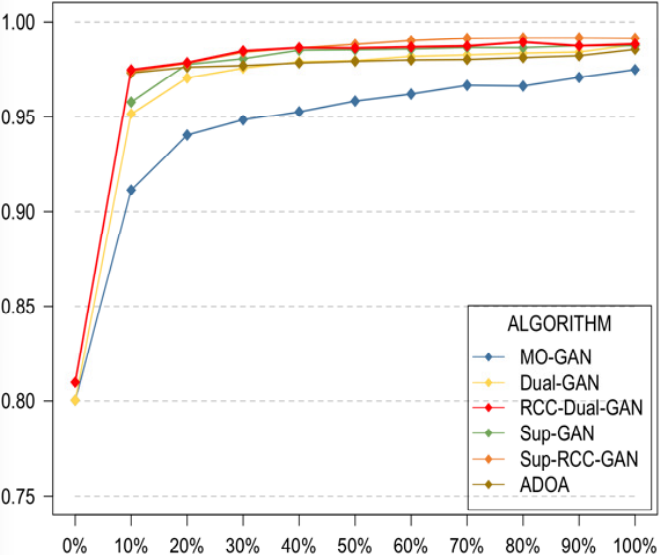}}
	\subfigure[\scriptsize{Har}]{
    	\label{fig:f8_10}
\includegraphics[width=0.19\textwidth]{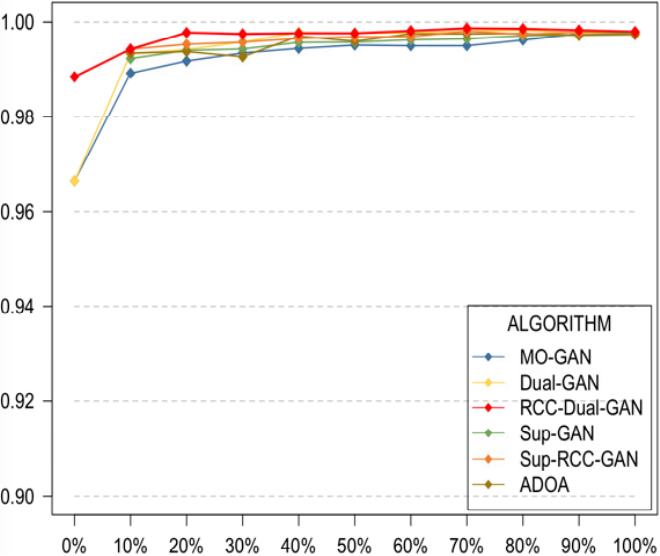}}
	\vspace{-5pt}
	\caption{Performance fluctuations of outlier detection algorithms on real-world datasets with different identification ratios.}
	\vspace{-15pt}
	\label{fig:f8}
\end{figure*}

{
	\begin{table*}[htb]
		\caption{Experimental Results of Outlier Detection Algorithms on Real-World Datasets}
		\vspace{-15pt}
		\label{tab:t_2}
		\centering
		\begin{tabular}[t]{p{1.8cm}p{0.8cm}p{0.8cm}p{1.1cm}p{1.3cm}p{1.4cm}p{2.2cm}p{2cm}p{1.6cm}p{0.8cm}}
        
			\hline
            Dataset & $k$NN & LOF & $k$-means & MO-GAN & Dual-GAN & RCC-Dual-GAN & Sup-RCC-GAN & Sup-GAN & ADOA\\
			\hline
            Thyroid  & 0.9365 & 0.9527 & 0.9381 & 0.9606 & 0.9775 & 0.9915 & \textbf{0.9970} & 0.9872 & 0.9927\\
            Pima & 0.7385 & 0.7154 & 0.6927 & 0.7366 & 0.7326 & \textbf{0.7460} & 0.6769 & 0.5823 & 0.7021\\
            Stamps & 0.9223 & 0.9010 & 0.9077 & 0.9236 & 0.9906 & \textbf{0.9922} & 0.9097 & 0.9022 & 0.9509\\
            Pageblocks & 0.8866 & 0.9232 & 0.9195 & 0.8456 & 0.9024 & \textbf{0.9317} & 0.9230 & 0.8066 & 0.9156\\
            Cardio & 0.9606 & 0.9639 & 0.9577 & 0.9516 & 0.9871 & \textbf{0.9892} & 0.9891 & 0.9105 & 0.9675\\
            Waveform & 0.8102 & 0.8071 & 0.7275 & 0.8658 & 0.9140 & 0.9184 & \textbf{0.9186} & 0.8821 & 0.8474\\
            Spambase & 0.5724 & 0.5391 & 0.5972 & 0.8947 & \textbf{0.9152} & 0.8785 & 0.9131 & 0.8753 & 0.8108\\
            Optdigits & 0.8303 & 0.9100 & 0.8843 & 0.9020 & 0.9926 & 0.9941 & 0.9960 & 0.9959 & \textbf{1.0000}\\
            Mnist & 0.8647 & 0.8562 & 0.8467 & 0.9114 & 0.9517 & \textbf{0.9748} & 0.9738 & 0.9579 & 0.9731\\
            Har & 0.9756 & 0.9827 & 0.9718 & 0.9892 & 0.9933 & \textbf{0.9943} & \textbf{0.9943} & 0.9923 & 0.9915\\
            \hline
            Average Rank & 6.9 & 6.0 & 7.6 & 5.7 & 3.7 & \textbf{1.9} & 2.9 & 6.0 & 4.2\\
			\hline
		\end{tabular}
		\vspace{-10pt}
	\end{table*}	
}
Compared with unsupervised methods (\ie $k$NN, LOF, and $k$-means), algorithms that use identified anomalies achieve substantially higher accuracy on most datasets, showing that reasonable use of these limited tags can effectively improve the performance of outlier detection even with only few identified anomalies. Moreover, to further evaluate the effect of the number of identified anomalies on different algorithms, semi-supervised and supervised approaches are performed on these datasets with different identification ratios. The results are shown in Fig. 8, where the ratio of identified anomalies in each dataset is adjusted from 0\% to 100\%. The accuracy of MO-GAN (shown with blue lines in Fig. \ref{fig:f8}) generally increases linearly with the identification ratio, and satisfactory results can only be obtained if there are many identified anomalies. By contrast, Dual-GAN, RCC-Dual-GAN, and Sup-RCC-GAN (shown with yellow, red, and orange line, respectively, in Fig. \ref{fig:f8}) can utilize few identified anomalies (\ie 10\% identification ratio) to achieve excellent results that approach the results when all tags are known (\ie 100\% identification ratio) on multiple datasets.

Compared with supervised methods, the overall performance (\ie average ranks) of Dual-GAN and RCC-Dual-GAN is superior to that of Sup-GAN and Sup-RCC-GAN, respectively. Although the suboptimal Sup-RCC-GAN achieves the best performance on three datasets (\ie Thyroid, Waveform, and Har), the identified anomalies in these datasets belong to one cluster (\ie NoC.=1). This means that all anomalies in each dataset are most likely generated by the same generation mechanism, and identified anomalies may represent all of them. If unidentified and emerging anomalies exist in the later detection, the accuracy of the supervised detector may not always be guaranteed. By contrast, the proposed semi-supervised methods, which also use the unsupervised modules (\ie UMO-GAN and UM-GAN) to establish the patterns of normal data, can simultaneously detect the partially identified group anomalies and all discrete anomalies.

The semi-supervised ADOA, which uses isolation and similarity to calculate the confidence of each instance, can identify partially identified group anomalies and discrete anomalies in the training data. However, due to the significant challenge that ADOA faces in detecting emerging anomalies, the overall performance of Dual-GAN and RCC-Dual-GAN is better than ADOA. Regarding the comparison between the two proposed methods, RCC-Dual-GAN outperforms Dual-GAN on nine of the ten datasets. It shows that the network structure combining RCC and M-GAN has greater stability in various datasets, which can also be reflected from the comparison between Sup-GAN and Sup-RCC-GAN.

\subsection{Applications}
\label{sec:4.2}
\subsubsection{Credit Card Fraud Detection}
\label{sec:4.2.1}
With the fast development of e-commerce, increasingly more kinds of credit card frauds arise, which poses a serious threat to all organizations issuing credit cards or managing online transactions. Thus, many machine learning and computational intelligence techniques have been proposed to reduce economic losses and simultaneously enhance customer confidence. However, they are mainly focused on the supervised or unsupervised setting, ignoring the verifiability of fraud and verification latency. That is, a small set of frauds can be timely checked by the investigator, whereas the remaining transactions will be unlabeled until customers discover fraud. Therefore, we apply our proposed models to the issue of credit card fraud detection.

Since banks are reluctant to disclose such data, we perform the experiment on a publicly available Credit-card dataset~\cite{Pozzolo2015Calibrating}. The Credit-card dataset contains 284,807 credit card records that occurred in two days of September 2013, where 492 records are fraudulent transactions. Each record consists of transaction time, amount, class (\ie normal or fraud) and 28 numerical features, which are the principal components extracted from the original features. On this basis, we further remove the transaction time and rescale the other features in the interval [0, 1]. And then, we divide the dataset into two datasets in the ratio of 2 to 1. The training dataset contains 328 fraudulent transactions out of 189,871 records, while the test dataset contains 164 fraudulent transactions out of 94,936 records. Finally, to match the special semi-supervised setting, we randomly select 10\% of fraudulent transactions (\ie 33 frauds) from the training dataset as identified frauds, and the remaining records are used as unlabeled transactions.

\begin{figure}[ht]
	\centering
	\includegraphics[width=0.42\textwidth]{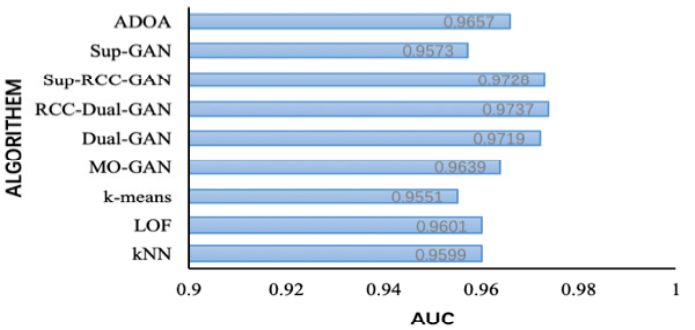}
	\vspace{-10pt}
	\caption{Experimental results on the Credit-card dataset.}
    \vspace{-10pt}
    \label{fig:f9}
\end{figure}

Experimental results on the Credit-card dataset are shown in Fig. \ref{fig:f9}. Similar to the results on real-world datasets, RCC-Dual-GAN and Dual-GAN obtain good performance, which demonstrates the effectiveness of our proposed methods on credit card fraud detection. The supervised Sup-RCC-GAN yields a suboptimal result because the identified frauds may represent the vast majority of fraudulent transactions. However, the detection accuracy of supervised Sup-GAN is even worse than that of unsupervised $k$NN and LOF. It indicates that the single GAN cannot accurately learn multiple generation mechanisms simultaneously, which can further prove the performance advantages of the combination of RCC and M-GAN.

\subsubsection{Network Intrusion Detection}
\label{sec:4.2.2}
Cybersecurity is another important application area for outlier detection, and a considerable number of machine learning techniques, including cluster-based, classification-based, and hybrid methods, have been developed for intrusion detection. However, although only part of intrusions can be detected in practice, semi-supervised methods are still rarely studied and applied to this issue, as discussed above. Thus, in this section, we apply our proposed methods to NSL-KDD, which is one of the most widely used datasets for performance evaluation of intrusion detection.

NSL-KDD solves several inherent problems of the KDD'99 by removing redundant records and readjusting its size. And then, in order to more suitable for the inherent nature that attacks are relatively uncommon, we further adjust the proportion of attacks by deleting 90\% of the attack records. Thus, the training dataset contains 67,343 normal records and 5,872 attacks, which belong to 21 attack types in four main categories (\ie DoS, Probe, R2L, and U2R); the test set contains 9,711 normal records and 1,304 attacks, which fall into 37 attack types. Finally, we randomly select 10\% of network intrusions (\ie 597 attacks in the 21 attack types) from each attack type in the training data as the identified attacks, and the remaining records are used as unlabeled behaviors.

\begin{figure}[ht]
	\centering
	\includegraphics[width=0.42\textwidth]{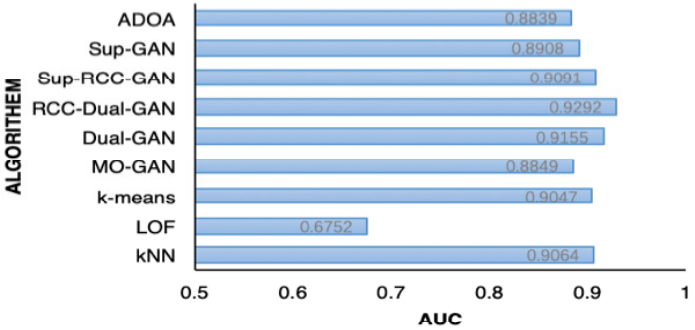}
	\vspace{-10pt}
	\caption{Experimental results on the NSL-KDD dataset.}
    \vspace{-10pt}
    \label{fig:f10}
\end{figure}

The experimental results on the NSL-KDD dataset are shown in Fig. \ref{fig:f10}. The semi-supervised RCC-Dual-GAN and Dual-GAN achieve the optimal and suboptimal outcomes, respectively, whereas the supervised Sup-GAN and Sup-RCC-GAN obtain results similar to the unsupervised $k$NN and $k$-means. This is most likely because only 19 of the 37 attack types in the test data are identified, so that the detection of emerging intrusions is as important as the effective use of identified attacks. As for the semi-supervised RCC-Dual-GAN and Dual-GAN, they can exploit the potential information in identified intrusions and simultaneously detect emerging discrete attacks.

\section{Conclusions and Future Works}
\label{sec:conclusions}

In this paper, we first propose a one-step method Dual-GAN for semi-supervised outlier detection with few identified anomalies, which can directly utilize the potential information in identified anomalies to detect partially identified group anomalies. In addition, since instances with similar output values may not all be similar in a complex data structure, we propose a modified model RCC-Dual-GAN based on Dual-GAN to create the reference distribution and augment the minority class more robustly. Considering the difficulty in finding the Nash equilibrium and optimal model during iteration, two evaluation indicators (\ie $NNR$ and $AP$) are provided to make the detection process more intelligent and reliable. Extensive experiments on synthetic data and real-world data show that even with only a few identified anomalies, our proposed approaches can substantially improve the accuracy of outlier detection. Moreover, credit card fraud detection and network intrusion detection are performed to demonstrate the effectiveness of our proposed methods in complex practical situations. In future, we attempt to introduce incremental learning into the training process to continuously learn new knowledge with less computational cost, and more intensive research on the evaluation of Nash equilibrium will be conducted.

\ifCLASSOPTIONcompsoc
  \section*{Acknowledgments}
\else
  \section*{Acknowledgment}
\fi
This work is supported by the Major Program of the National Natural Science Foundation of China (91846201, 71490725), the Foundation for Innovative Research Groups of the National Natural Science Foundation of China (71521001), the National Natural Science Foundation of China (71722010, 91546114, 91746302, 71872060), The National Key Research and Development Program of China (2017YFB0803303).

\ifCLASSOPTIONcaptionsoff
  \newpage
\fi

\bibliographystyle{IEEEtran}
\bibliography{reference.bib}

\begin{thebibliography}{10}
\providecommand{\url}[1]{#1}
\csname url@samestyle\endcsname
\providecommand{\newblock}{\relax}
\providecommand{\bibinfo}[2]{#2}
\providecommand{\BIBentrySTDinterwordspacing}{\spaceskip=0pt\relax}
\providecommand{\BIBentryALTinterwordstretchfactor}{4}
\providecommand{\BIBentryALTinterwordspacing}{\spaceskip=\fontdimen2\font plus
\BIBentryALTinterwordstretchfactor\fontdimen3\font minus
  \fontdimen4\font\relax}
\providecommand{\BIBforeignlanguage}[2]{{%
\expandafter\ifx\csname l@#1\endcsname\relax
\typeout{** WARNING: IEEEtran.bst: No hyphenation pattern has been}%
\typeout{** loaded for the language `#1'. Using the pattern for}%
\typeout{** the default language instead.}%
\else
\language=\csname l@#1\endcsname
\fi
#2}}
\providecommand{\BIBdecl}{\relax}
\BIBdecl

\bibitem{Fiore2017Using}
U.~Fiore, A.~D. Santis, F.~Perla, P.~Zanetti, and F.~Palmieri, ``Using
  generative adversarial networks for improving classification effectiveness in
  credit card fraud detection,'' \emph{Information Sciences}, 2017.

\bibitem{Pozzolo2017Credit}
A.~D. Pozzolo, G.~Boracchi, O.~Caelen, C.~Alippi, and G.~Bontempi, ``Credit
  card fraud detection: A realistic modeling and a novel learning strategy,''
  \emph{IEEE Transactions on Neural Networks and Learning Systems}, vol.~29,
  no.~8, pp. 3784--3797, 2017.

\bibitem{Makki2019An}
S.~Makki, Z.~Assaghir, Y.~Taher, R.~Haque, M.-S. Hacid, and H.~Zeineddine, ``An
  experimental study with imbalanced classification approaches for credit card
  fraud detection,'' \emph{IEEE Access}, vol.~7, pp. 93\,010--93\,022, 2019.

\bibitem{Rahman2017Search}
M.~Rahman, M.~Rahman, B.~Carbunar, and D.~H. Chau, ``Search rank fraud and
  malware detection in google play,'' \emph{IEEE Transactions on Knowledge and
  Data Engineering}, vol.~29, no.~8, pp. 1329--1342, 2017.

\bibitem{Liu2019A}
S.~Liu, B.~Hooi, and C.~Faloutsos, ``A contrast metric for fraud detection in
  rich graphs,'' \emph{IEEE Transactions on Knowledge and Data Engineering},
  vol.~31, no.~12, pp. 2235--2248, 2019.

\bibitem{Preeti2019A}
P.~Mishra, V.~Varadharajan, U.~Tupakula, and E.~S. Pilli, ``A detailed
  investigation and analysis of using machine learning techniques for intrusion
  detection,'' \emph{IEEE Communications Surveys and Tutorials}, vol.~21,
  no.~1, pp. 686--728, 2019.

\bibitem{Raman2017A}
M.~R.~G. Raman, N.~Somu, K.~Kirthivasan, and V.~S. Sriram, ``A hypergraph and
  arithmetic residue-based probabilistic neural network for classification in
  intrusion detection systems,'' \emph{Neural Networks}, vol.~92, pp. 89--97,
  2017.

\bibitem{Mao2017Feature}
J.~Mao, T.~Wang, C.~Jin, and A.~Zhou, ``Feature grouping-based outlier
  detection upon streaming trajectories,'' \emph{IEEE Transactions on Knowledge
  and Data Engineering}, vol.~29, no.~12, pp. 2696--2709, 2017.

\bibitem{Yang2009Outlier}
X.~Yang, L.~J. Latecki, and D.~Pokrajac, ``Outlier detection with globally
  optimal exemplar-based gmm,'' in \emph{SIAM International Conference on Data
  Mining}, 2009, pp. 145--154.

\bibitem{Bo2018Deep}
B.~Zong, Q.~Song, M.~R. Min, W.~Cheng, C.~Lumezanu, D.~Cho, and H.~Chen, ``Deep
  autoencoding gaussian mixture model for unsupervised anomaly detection,'' in
  \emph{International Conference on Learning Representations}, 2018.

\bibitem{Manzoor2016Fast}
E.~Manzoor, S.~M. Milajerdi, and L.~Akoglu, ``Fast memory-efficient anomaly
  detection in streaming heterogeneous graphs,'' in \emph{ACM SIGKDD
  International Conference on Knowledge Discovery and Data Mining}, 2016, pp.
  1035--1044.

\bibitem{Paulheim2015A}
H.~Paulheim and R.~Meusel, ``A decomposition of the outlier detection problem
  into a set of supervised learning problems,'' \emph{Machine Learning}, vol.
  100, no. 2-3, pp. 509--531, 2015.

\bibitem{Salehi2016Fast}
M.~Salehi, C.~Leckie, J.~C. Bezdek, T.~Vaithianathan, and X.~Zhang, ``Fast
  memory efficient local outlier detection in data streams,'' \emph{IEEE
  Transactions on Knowledge and Data Engineering}, vol.~28, no.~12, pp.
  3246--3260, 2016.

\bibitem{Chehreghani2016K}
M.~H. Chehreghani, ``K-nearest neighbor search and outlier detection via
  minimax distances,'' in \emph{SIAM International Conference on Data Mining},
  2016, pp. 405--413.

\bibitem{Makki2019Adapted}
Y.~Djenouri, A.~Belhadi, J.~C.-W. Lin, and A.~Cano, ``Adapted k-nearest
  neighbors for detecting anomalies on spatio–temporal traffic flow,''
  \emph{IEEE Access}, vol.~7, pp. 10\,015--10\,027, 2019.

\bibitem{Zhou2017Anomaly}
C.~Zhou and R.~C. Paffenroth, ``Anomaly detection with robust deep
  autoencoders,'' in \emph{ACM SIGKDD International Conference on Knowledge
  Discovery and Data Mining}, 2017, pp. 665--674.

\bibitem{Schlegl2017Unsupervised}
T.~Schlegl, P.~Seeböck, S.~M. Waldstein, U.~Schmidt-Erfurth, and G.~Langs,
  ``Unsupervised anomaly detection with generative adversarial networks to
  guide marker discovery,'' in \emph{International Conference on Information
  Processing in Medical Imaging}, 2017, pp. 146--157.

\bibitem{Adversarially2018Sabokrou}
M.~Sabokrou, M.~Khalooei, M.~Fathy, and E.~Adeli, ``Adversarially learned
  one-class classifier for novelty detection,'' in \emph{IEEE Conference on
  Computer Vision and Pattern Recognition}, 2018, p. 3379–3388.

\bibitem{Steinwart2005A}
I.~Steinwart, ``A classification framework for anomaly detection,''
  \emph{Journal of Machine Learning Research}, vol.~6, no.~1, pp. 211--232,
  2005.

\bibitem{Zhang2018Anomaly}
Y.~Zhang, L.~Li, J.~Zhou, X.~Li, and Z.~Zhou, ``Anomaly detection with
  partially observed anomalies,'' in \emph{WWW: International World Wide Web
  Conference}, 2018, pp. 639--646.

\bibitem{Aggarwal2017Outlier}
C.~C. Aggarwal, \emph{Outlier Analysis}.\hskip 1em plus 0.5em minus 0.4em\relax
  Springer International Publishing, 2017.

\bibitem{Daneshpazhouh2014Entropy}
A.~Daneshpazhouh and A.~Sami, ``Entropy-based outlier detection using
  semi-supervised approach with few positive examples,'' \emph{Pattern
  Recognition Letters}, vol.~49, pp. 77--84, 2014.

\bibitem{Daneshpazhouh2014Semi}
------, ``Semi-supervised outlier detection with only positive and unlabeled
  data based on fuzzy clustering,'' \emph{International Journal on Artificial
  Intelligence Tools}, vol.~24, no.~3, 2015.

\bibitem{Bo2014An}
B.~Liu, Y.~Xiao, P.~S. Yu, Z.~Hao, and L.~Cao, ``An efficient approach for
  outlier detection with imperfect data labels,'' \emph{IEEE Transactions on
  Knowledge and Data Engineering}, vol.~26, no.~7, pp. 1602--1616, 2014.

\bibitem{Liu2019Generative}
Y.~Liu, Z.~Li, C.~Zhou, Y.~Jiang, J.~Sun, M.~Wang, and X.~He, ``Generative
  adversarial active learning for unsupervised outlier detection,'' \emph{IEEE
  Transactions on Knowledge and Data Engineering}, 2019.

\bibitem{Shah2017Robust}
S.~A. Shah and V.~Koltun, ``Robust continuous clustering,'' \emph{Proceedings
  of the National Academy of Sciences}, vol. 114, no.~37, p. 9814–9819, 2017.

\bibitem{Wang2019Progress}
H.~Wang, M.~Bah, and M.~Hammad, ``Progress in outlier detection techniques: A
  survey,'' \emph{IEEE Access}, vol.~7, pp. 107\,964--108\,000, 2019.

\bibitem{Wang2010Boosting}
B.~X. Wang and N.~Japkowicz, ``Boosting support vector machines for imbalanced
  data sets,'' \emph{Knowledge and Information Systems}, vol.~25, no.~1, pp.
  1--20, 2010.

\bibitem{Lima2017Feature}
R.~F. Lima and A.~C.~M. Pereira, ``Feature selection approaches to fraud
  detection in e-payment systems,'' in \emph{International Conference on
  Electronic Commerce and Web Technologies}, 2017, pp. 111--126.

\bibitem{Lima2019Heartbeat}
J.~L.~P. Lima, D.~Macêdo, and C.~Zanchettin, ``Heartbeat anomaly detection
  using adversarial oversampling,'' in \emph{IEEE International Joint
  Conference on Neural Networks}, 2019.

\bibitem{Erfani2016High}
S.~M. Erfani, S.~Rajasegarar, S.~Karunasekera, and C.~Leckie,
  ``High-dimensional and large-scale anomaly detection using a linear one-class
  svm with deep learning,'' \emph{Pattern Recognition}, vol.~58, pp. 128--134,
  2016.

\bibitem{Liu2013SVDD}
B.~Liu, Y.~Xiao, L.~Cao, Z.~Hao, and F.~Deng, ``Svdd-based outlier detection on
  uncertain data,'' \emph{Knowledge and Information Systems}, vol.~34, no.~3,
  pp. 597--618, 2013.

\bibitem{Gao2006Semi}
J.~Gao, H.~Cheng, and P.~Tan, ``Semi-supervised outlier detection,'' in
  \emph{ACM symposium on Applied computing}, 2006, pp. 635--636.

\bibitem{Xue2010Semi}
Z.~Xue, Y.~Shang, and A.~Feng, ``Semi-supervised outlier detection based on
  fuzzy rough c-means clustering,'' \emph{Mathematics and Computers in
  Simulation}, vol.~80, no.~9, pp. 1911--1921, 2010.

\bibitem{Goodfellow2014Generative}
I.~J. Goodfellow, J.~Pouget-Abadie, M.~Mirza, B.~Xu, D.~Warde-Farley, S.~Ozair,
  A.~Courville, and Y.~Bengio, ``Generative adversarial networks,''
  \emph{Advances in Neural Information Processing Systems}, vol.~3, pp.
  2672--2680, 2014.

\bibitem{Akcay2018GANomaly}
S.~Akcay, A.~Atapourabarghouei, and T.~P. Breckon, ``Ganomaly: Semi-supervised
  anomaly detection via adversarial training,'' \emph{arXiv:1805.06725}, 2018.

\bibitem{Zenati2018Efficient}
H.~Zenati, C.~S. Foo, B.~Lecouat, G.~Manek, and V.~R. Chandrasekhar,
  ``Efficient gan-based anomaly detection,'' in \emph{The Workshop on
  International Conference on Learning Representations}, 2018.

\bibitem{Bian2019A}
J.~Bian, X.~Hui, S.~Sun, X.~Zhao, and M.~Tan, ``A novel and efficient
  cvae-gan-based approach with informative manifold for semi-supervised anomaly
  detection,'' \emph{IEEE Access}, vol.~7, pp. 88\,903--88\,916, 2019.

\bibitem{Wang2018Anomaly}
C.~Wang, Y.~Zhang, and C.~Liu, ``Anomaly detection via minimum likelihood
  generative adversarial networks,'' in \emph{International Conference on
  Pattern Recognition}, 2018.

\bibitem{Swee2018DOPING}
S.~K. Lim, Y.~Loo, N.-T. Tran, N.-M. Cheung, G.~Roig, and Y.~Elovici, ``Doping:
  Generative data augmentation for unsupervised anomaly detection with gan,''
  in \emph{IEEE International Conference on Data Mining}, 2018.

\bibitem{Zheng2018Generative}
Y.~J. Zheng, X.~Zhou, W.~Sheng, Y.~Xue, and S.~Chen, ``Generative adversarial
  network based telecom fraud detection at the receiving bank,'' \emph{Neural
  Networks}, vol. 102, pp. 78--86, 2018.

\bibitem{Kimura2018Semi}
M.~Kimura and T.~Yanagihara, ``Semi-supervised anomaly detection using gans for
  visual inspection in noisy training data,'' \emph{arXiv:1807.01136}, 2018.

\bibitem{Gao2018Low}
H.~Gao, Z.~Shou, A.~Zareian, H.~Zhang, and S.~Chang, ``Low-shot learning via
  covariance-preserving adversarial augmentation networks,'' in \emph{Advances
  in Neural Information Processing Systems}, 2018, pp. 981--991.

\bibitem{Campos2016On}
G.~O. Campos, A.~Zimek, J.~Sander, R.~J. G.~B. Campello, B.~Micenková,
  E.~Schubert, I.~Assent, and M.~E. Houle, ``On the evaluation of unsupervised
  outlier detection: measures, datasets, and an empirical study,'' \emph{Data
  Mining and Knowledge Discovery}, vol.~30, no.~4, pp. 891--927, 2016.

\bibitem{Pozzolo2015Calibrating}
A.~D. Pozzolo, O.~Caelen, R.~A. Johnson, and G.~Bontempi, ``Calibrating
  probability with undersampling for unbalanced classification,'' in \emph{IEEE
  Symposium Series on Computational Intelligence}, 2015, pp. 159--166.

\end{thebibliography}

\vspace{-20pt}
\begin{IEEEbiography}[{\includegraphics[width=1in,height=1.25in,clip,keepaspectratio]{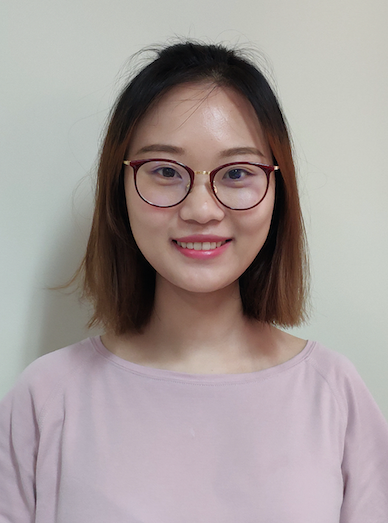}}]{Zhe Li} is working toward the PhD degree in Management Science and Engineering from Hefei University of Technology. She received her BE degree in Information Management and Information System from Hefei University of Technology, Hefei, China. Her research interests include machine learning and data mining, especially outlier detection, class imbalance learning, and ensemble learning.
\end{IEEEbiography}
\vspace{-20pt}
\begin{IEEEbiography}
[{\includegraphics[width=1in,height=1.25in,clip,keepaspectratio]{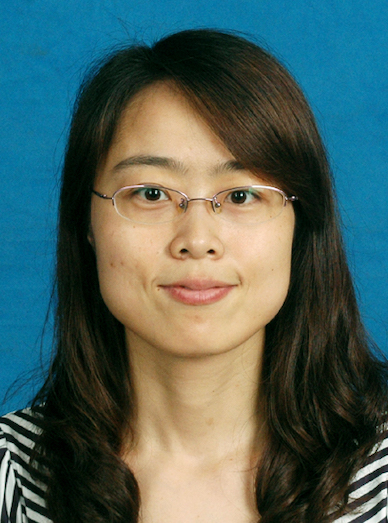}}]{Chunhua Sun}is an associate professor of Electronic Commerce at the Hefei University of Technology. She received his PhD in Management Science and Engineering from Hefei University of Technology, Hefei, China. Her research interests include electronic commerce, online consumer behavior and data mining. She has published papers in journals such as Scientific Reports, Discrete Dynamics in Nature, Society and Future Generation Computer Systems and IEEE Transactions on Professional Communication.
\end{IEEEbiography}
\vspace{-10pt}
\begin{IEEEbiography}
[{\includegraphics[width=1in,height=1.25in,clip,keepaspectratio]{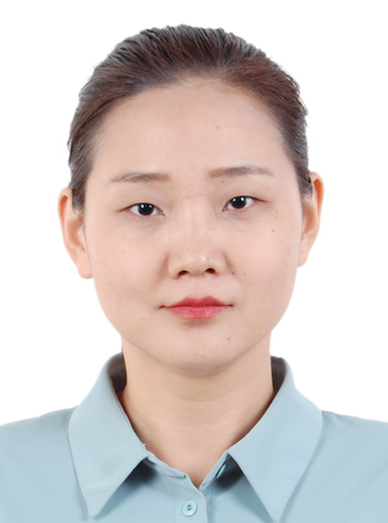}}]{Chunli Liu}is an assistant professor in School of Management, Hefei University of Technology, China. She received the PhD degree in management from Sun Yat-sen University. Her current research interests include electronic commerce, financial technology, data mining and financial fraud detection.
\end{IEEEbiography}
\vspace{-10pt}
\begin{IEEEbiography}
[{\includegraphics[width=1in,height=1.25in,clip,keepaspectratio]{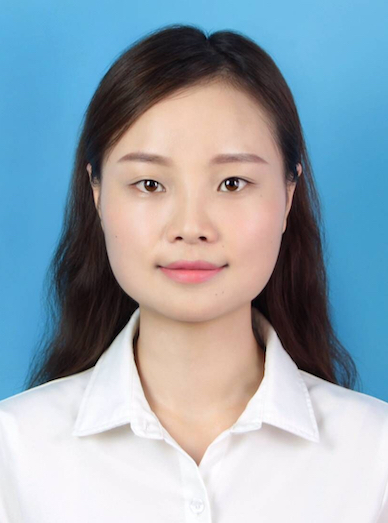}}]{Xiayu Chen}is an instructor of Electronic Commerce at the Hefei University of Technology. She received his PhD in Information Systems from the University of Science and Technology of China and the City University of Hong Kong. Her research interests include electronic commerce, social media and data mining. She has published papers in journals such as Information Systems Journal, International Journal of Electronic Commerce, Journal of Information Technology, Information Technology and People, International Journal of Information Management and Computers in Human Behavior.
\end{IEEEbiography}
\vspace{-10pt}
\begin{IEEEbiography}[{\includegraphics[width=1in,height=1.25in,clip,keepaspectratio]{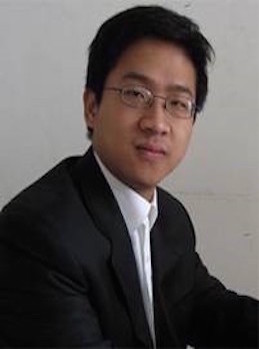}}]{Meng Wang}is a professor with the Hefei University of Technology. He received his BE and PhD degrees from the Special Class for the Gifted Young, Department of Electronic Engineering and Information Science, University of Science and Technology of China (USTC), Hefei, China, in 2003 and 2008, respectively. His current research interests include multimedia content analysis, computer vision, and pattern recognition. He has authored more than 200 book chapters, and journal and conference papers in these areas. He received the ACM SIGMM Rising Star Award 2014. He is an associate editor of the IEEE Transactions on Knowledge and Data Engineering and the IEEE Transactions on Circuits and Systems for Video Technology. He is a member of the IEEE and ACM.
\end{IEEEbiography}
\vspace{-10pt}
\begin{IEEEbiography}
[{\includegraphics[width=1in,height=1.25in,clip,keepaspectratio]{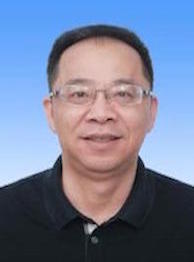}}]{Yezheng Liu}is a professor of Electronic Commerce at Hefei University of Technology. He received his PhD in Management Science and Engineering from Hefei University of Technology in 2001. His main research interests include data mining, decision science, electronic commerce, and intelligent decision support systems. His current research focuses on big data analytics, online social network, personalized recommendation system and outlier detection. He is the author and coauthor of numerous papers in scholarly journals, including IEEE Transactions on Knowledge and Data Engineering, Marketing Science, Decision Support Systems, International Journal of Production Economics, Knowledge-Based Systems, Journal of Management Sciences in China. He is a national member of New Century Talents Project.
\end{IEEEbiography}
\vfill
\end{document}